\documentclass[final]{cvpr}

\usepackage{comment}
\usepackage{times}
\usepackage{epsfig}
\usepackage{graphicx}
\usepackage{amsmath}
\usepackage{amssymb}
\usepackage{caption}
\usepackage{cite}
\usepackage{floatrow}
\usepackage[label font=normalfont]{subfig}
\usepackage[pagebackref=true,breaklinks=true,colorlinks,bookmarks=false]{hyperref}

\newcommand{\Section}[1]{\vspace{-1mm} \section{#1} \vspace{1mm}}
\newcommand{\SubSection}[1]{\vspace{-1mm} \subsection{#1} \vspace{-1mm}}

\newcommand{\Paragraph}[1]{\vspace{1.25mm}\noindent\textbf{#1.}\hspace{0.5mm}}

\def\cvprPaperID{6522} % *** Enter the CVPR Paper ID here
\def\confYear{CVPR 2021}

\begin{document}

%%%%%%%%% TITLE
\title{Towards High Fidelity Face Relighting with Realistic Shadows}

% For a paper whose authors are all at the same institution,
% omit the following lines up until the closing ``}''.
% Additional authors and addresses can be added with ``\and'',
% just like the second author.
% To save space, use either the email address or home page, not both

\author{Andrew Hou\thanks{All of the data mentioned in this paper was downloaded and used at Michigan State University. } $^{,1}$, Ze Zhang$^{1}$, Michel Sarkis$^{2}$, Ning Bi$^{2}$, Yiying Tong$^{1}$, Xiaoming Liu$^{1}$ \\
{ $^{1}$Michigan State University, $^{2}$Qualcomm Technologies Inc.
} \\
{\tt\small \{houandr1, zhangze6, ytong, liuxm\}@msu.edu, \{msarkis, nbi\}@qti.qualcomm.com} \\
{\small \url{https://github.com/andrewhou1/Shadow-Mask-Face-Relighting}}
}

\makeatletter
\let\@oldmaketitle\@maketitle% Store \@maketitle
\renewcommand{\@maketitle}{\@oldmaketitle% Update \@maketitle to insert...
\vspace{-9mm}
\begin{center}
\begin{minipage}[t]{0.16\linewidth}
\centering
\includegraphics[width=\linewidth]{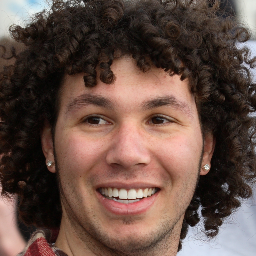}
\end{minipage}
\begin{minipage}[t]{0.16\linewidth}
\centering
\includegraphics[width=\linewidth]{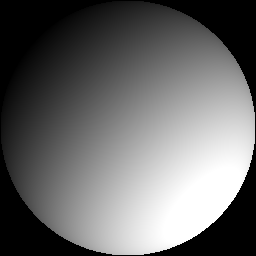}
\end{minipage}
\begin{minipage}[t]{0.16\linewidth}
\centering
\includegraphics[width=\linewidth]{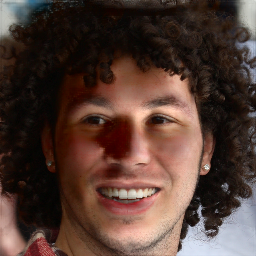}
\end{minipage}
\begin{minipage}[t]{0.16\linewidth}
\centering
\includegraphics[width=\linewidth]{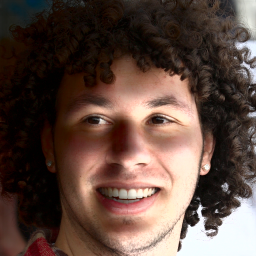}
\end{minipage}
\begin{minipage}[t]{0.16\linewidth}
\centering
\includegraphics[width=\linewidth]{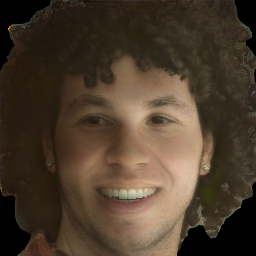}
\end{minipage}
\begin{minipage}[t]{0.16\linewidth}
\centering
\includegraphics[width=\linewidth]{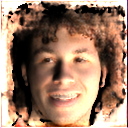}
\end{minipage}

%\newline

\begin{minipage}[t]{0.16\linewidth}
\centering
\includegraphics[width=\linewidth]{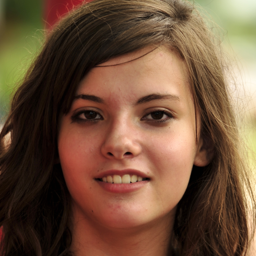} \\
\small (a) Source image
\end{minipage}
\begin{minipage}[t]{0.16\linewidth}
\centering
\includegraphics[width=\linewidth]{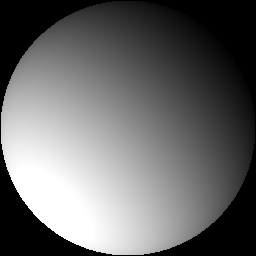} \\
\small (b) Target lighting
\end{minipage}
\begin{minipage}[t]{0.16\linewidth}
\centering
\includegraphics[width=\linewidth]{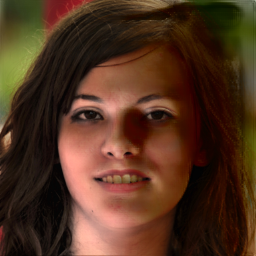} \\
\small (c) Our model
\end{minipage}
\begin{minipage}[t]{0.16\linewidth}
\centering
\includegraphics[width=\linewidth]{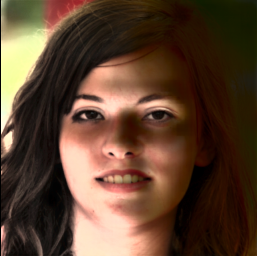} \\
\small (d) DPR \cite{DPR}
\end{minipage}
\begin{minipage}[t]{0.16\linewidth}
\centering
\includegraphics[width=\linewidth]{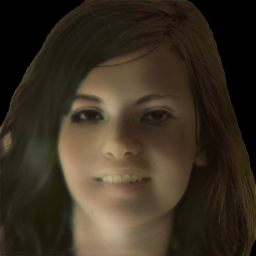} \\
\small (e) SIPR \cite{UCSDSingleImagePortraitRelighting}
\end{minipage}
\begin{minipage}[t]{0.16\linewidth}
\centering
\includegraphics[width=\linewidth]{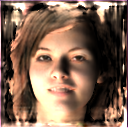} \\
\small (f) SfSNet \cite{SfSNet}
\end{minipage}
\end{center}
\vspace{-2mm}
{\small Figure $1$:  Given an input image and a target lighting, our model can produce a relit image with realistic hard cast shadows, such as those around the nose and eyes, which prior work struggles with. Images of SIPR~\cite{UCSDSingleImagePortraitRelighting} are provided by the authors.}
 \bigskip}% ... an image
\makeatother

\renewcommand{\figurename}{Figure}

\maketitle
\thispagestyle{empty}
\pagestyle{empty}

%%%%%%%%% ABSTRACT
\begin{abstract}
   \vspace{-4mm}
   Existing face relighting methods often struggle with two problems: maintaining the local facial details of the subject and accurately removing and synthesizing shadows in the relit image, especially hard shadows. We propose a novel deep face relighting method that addresses both problems. Our method learns to predict the ratio (quotient) image between a source image and the target image with the desired lighting, allowing us to relight the image while maintaining the local facial details. During training, our model also learns to accurately modify shadows by using estimated shadow masks to emphasize on the high-contrast shadow borders. Furthermore, we introduce a method to use the shadow mask to estimate the ambient light intensity in an image, and are thus able to leverage multiple datasets during training with different global lighting intensities. With quantitative and qualitative evaluations on the Multi-PIE and FFHQ datasets, we demonstrate that our proposed method faithfully maintains the local facial details of the subject and can accurately handle hard shadows while achieving state-of-the-art face relighting performance. 
\end{abstract}
\vspace{-3mm}

%%%%%%%%% BODY TEXT
\Section{Introduction}

Face relighting is the problem of turning a source image of a human face into a new image of the same face under a desired illumination different from the original lighting. It has long been studied in computer vision and computer graphics and has a wide range of applications in face-related problems such as face recognition~\cite{HaLeWACV19, facerecognition,improving-face-recognition-from-hard-samples-via-distribution-distillation-loss} and entertainment~\cite{PeersSIGGRAPH2007}. With the everlasting interest in consumer photography and photo editing, the ability to produce realistic relit face images will remain an important problem. 

Many existing face relighting models utilize intrinsic decomposition of the image into face geometry, lighting, and reflectance~\cite{SfSNet, PhysicsGuidedRelighting, WangPAMI2009, EggerIJCV2018, NeuralFaceEditing, HaLeWACV19, LuanCVPR18, FreemanIntrinsic2018, Tewari2017, YamaguchiIntrinsic2018, BarronMalikSfS, RealisticInverseLighting, Pfister, TowardsHighFidelityFaceReconstruction, UncertaintyFaceReconstruction, IntrinsicFacePriors}. The source image is then relit by rendering with a novel illumination. Other relighting methods employ image-to-image translation~\cite{UCSDSingleImagePortraitRelighting, DPR, PhysicsGuidedRelighting} or style transfer~\cite{Flickr, DeepPhotoStyleTransfer, ClosedFormSolution, MassTransport}. 

For most face relighting applications, one important requirement is the preservation of the subject's local facial details during relighting. Intrinsic decomposition methods often compromise high frequency details and can leave artifacts in the relit face images due to errors in the geometry or reflectance estimation. Another important feature of a desirable face relighting model is proper shadow handling. For entertainment in particular, adding and removing shadows accurately is crucial in producing photorealistic results. Most existing relighting methods, however, do not model {\it hard self-cast shadows} caused by directional lights. 

Our proposed method uses an hourglass network to formulate the relighting problem as a ratio (quotient) image~\cite{ShashuaRatioImage} estimation problem. 
The ratio image estimated by our model can be multiplied with the source image to generate the target image under the new illumination. This approach allows our model to maintain the local facial details of the subject while adjusting the intensity of each pixel during relighting. We employ an estimation loss to enable ratio image learning, as well as a structural dissimilarity (DSSIM) loss based on the SSIM metric~\cite{SSIM} to enhance the perceptual quality of relit faces. In addition, we incorporate PatchGAN~\cite{PatchGAN} to further improve the plausibility. 

During training, we generate and leverage shadow masks, which indicate estimated shadow regions for each image using the lighting direction and $3$D shape from $3$D Morphable Model ($3$DMM)~\cite{3DMM} fitting. The shadow masks enable us to handle shadows through {\it weighted} ratio image estimation losses. We place higher emphasis on the pixels close to shadow borders in the source and target relighting images, with larger weights placed on borders of high-contrast cast shadows over soft ones. 
This simple strategy allows learning how to accurately add and remove both hard and soft shadows under various relighting scenarios. 

Our training process can leverage  images with both diffuse and directional lighting across multiple datasets, which improves our ability to handle diverse lighting and generalize to unseen data over methods that only train on a single %light stage (maybe not so explicit?)
dataset~\cite{UCSDSingleImagePortraitRelighting, PhysicsGuidedRelighting, DPR}. To enable this, we use our shadow masks to estimate the ambient lighting intensity in each image, and modify our lighting to account for differences in ambient lighting across images and datasets. Thus, our model accommodates for  differences in the environment between images in controlled and in-the-wild settings. 

%\noindent 
Our proposed method has three main contributions: 
%\begin{itemize}
%\setlength\itemsep{1mm}
  %\item
  
  % $\diamond$ We propose the first ratio-image based face relighting method that models the effects of strong directional lights while requiring only a single image and a target lighting as input. Our use of the ratio image allows better preservation of the local facial details of the relit subject than existing methods.
  $\diamond$ We propose a novel face relighting method that models both high-contrast cast shadows and soft shadows, while preserving the local facial details.
   
  $\diamond$ Our technical approach involves single image based ratio image estimation to better preserve local details, shadow border reweighting to handle hard shadows, and ambient light compensation to account for dataset differences.
  
  %improving shadow handling through shadow masks, by simply placing higher emphasis on learning the ratio image near shadow borders and high-contrast cast shadows.
  
  %$\diamond$ We introduce a single image ambient lighting estimation by using shadow masks. This allows our method to account for differences in the environment across different relighting datasets, and enriches our model by combining multiple datasets during training.
  
  $\diamond$ Our approach achieves the state-of-the-art relighting results  on two benchmarks quantitatively and qualitatively.
%\end{itemize}

%------------------------------------------------------------------------
\Section{Related Work}
%------------------------------------------------------------------------
\renewcommand{\thefigure}{2}
\begin{figure*}[t]
\vspace{-2mm}
\begin{center}
   \includegraphics[width=0.82 \linewidth]{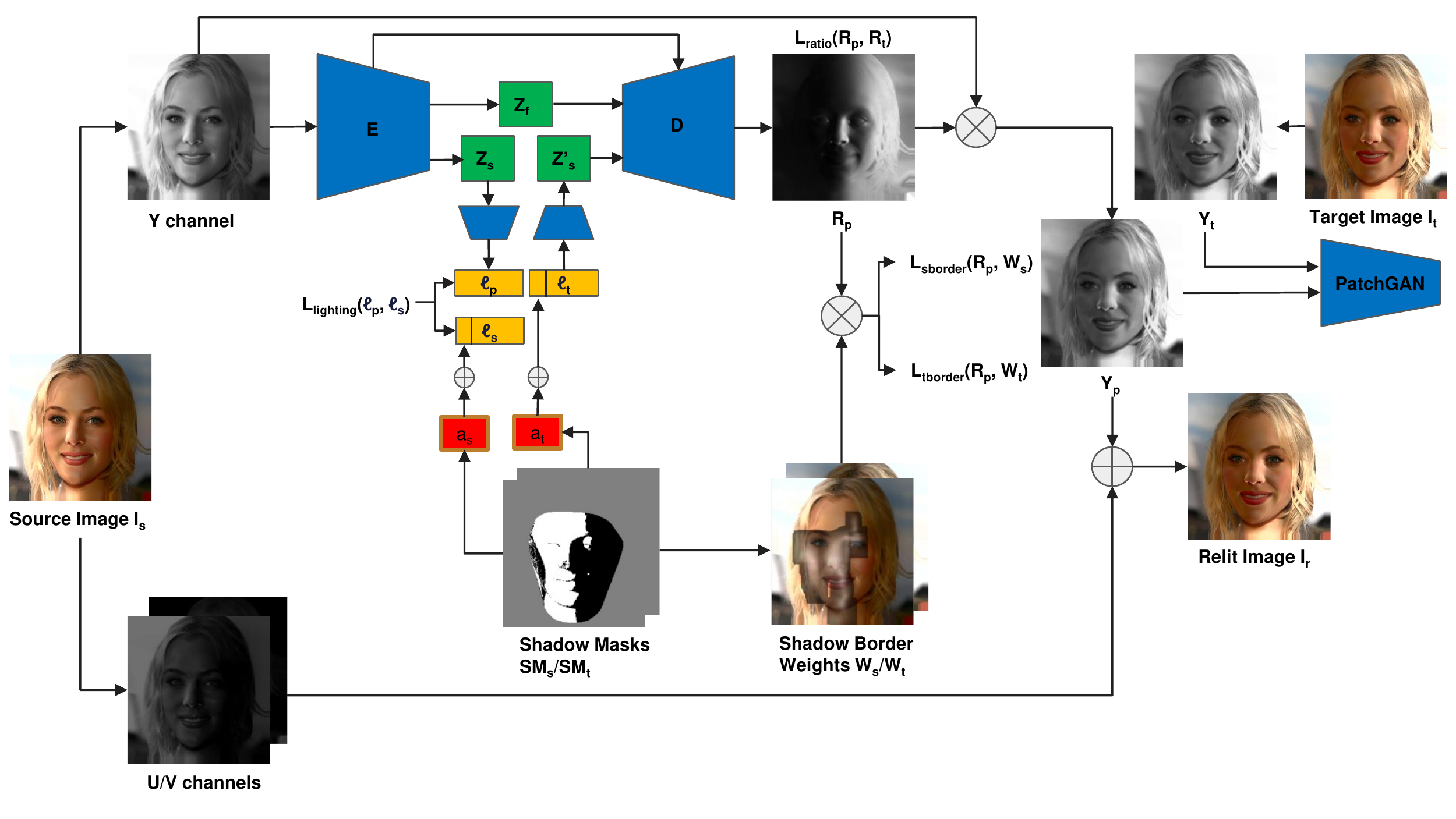}
\vspace{0mm}
\caption{\small \textbf{Overview of our proposed method}. Given a source image $\mathbf{I}_{s}$ and a target lighting $\mathbf{\ell}_{t}$ as input, our method estimates the source lighting $\mathbf{\ell}_{p}$ and the ratio image $\mathbf{R}_{p}$ in the luminance channel. In training, we assign high emphasis to learning the correct ratio image at pixels near shadow borders in both source and target images, helping the model add and remove shadows more accurately. We also estimate the ambient light intensity of the source image ($\mathbf{a}_{s}$) and of the target image ($\mathbf{a}_{t}$) and add them to the groundtruth source lighting $\mathbf{\ell}_{s}$ and target lighting $\mathbf{\ell}_{t}$ respectively to help our model adapt to images across different datasets with varying ambient light. 
}\label{fig:Architecture}
\end{center}\vspace{-8mm}
\end{figure*}
\Paragraph{Face Relighting}
Among prior face relighting works, many relight via intrinsic decomposition and rendering~\cite{SfSNet, PhysicsGuidedRelighting, WangPAMI2009, EggerIJCV2018, NeuralFaceEditing, HaLeWACV19, LuanCVPR18,towards-high-fidelity-nonlinear-3d-face-morphable-model, on-learning-3d-face-morphable-model-from-in-the-wild-images, FreemanIntrinsic2018, Tewari2017, YamaguchiIntrinsic2018, BarronMalikSfS, RealisticInverseLighting, Pfister, TowardsHighFidelityFaceReconstruction, UncertaintyFaceReconstruction, IntrinsicFacePriors}: the source image is decomposed into face geometry, reflectance, and lighting, and  render relit images via modified lighting. 
As decomposition generally relies heavily on single image face reconstruction, which remains an open problem, these methods tend to produce results that lack high frequency detail found in the source image and contain artifacts from geometry and reflectance estimation error. 
Our method bypasses this issue by avoiding intrinsic decomposition and estimating a ratio image instead. The ratio image only affects the intensity of each pixel in the source image in a smooth way aside from shadow borders, thus preserving the local facial details. % of the subject. 

Other relighting methods do not perform explicit intrinsic decomposition. % the source images.  into intrinsic components. 
Sun {\it et~al.}~\cite{UCSDSingleImagePortraitRelighting} use image-to-image translation to produce high quality relighting. % via environment maps. 
However, their results deteriorate when the input image contains hard shadows or sharp specularities, or when presented with strong directional light. %Deep Single-Image Portrait Relighting 
Zhou {\it et~al.}~\cite{DPR} use a Spherical Harmonics (SH) lighting model and estimate the target luminance from the source luminance and a target lighting. However, their model is trained on images with primarily diffuse lighting and only handles soft shadows. 
We train our model on a mixture of images with diffuse and strong directional light. We also assign larger emphasis on learning the ratio image near the shadow borders of high-contrast cast shadows. % using estimated shadow masks. 
Hence, our model handles shadows effectively. 

Some methods relight using a style transfer approach~\cite{Flickr, DeepPhotoStyleTransfer, ClosedFormSolution, MassTransport} by transferring the lighting conditions of a reference image as a style to a source image. However, since they require a reference image for each unique target lighting, they are less flexible to use in practice. Our model only requires a source image and a target lighting as input. 

More recently, Zhang \textit{et al.} \cite{PortraitShadowManipulation} propose a method that can remove both foreign and facial shadows and can be considered a relighting method. While their method is excellent in removing facial shadows, it tends to not synthesize realistic hard shadows during relighting. 

%------------------------------------------------------------------------
\Paragraph{Ratio Images in Face Relighting}
%Ratio images for face relighting was first proposed by Shashua and Riklin-Raviv~\cite{ShashuaRatioImage}. 
Prior face relighting works that incorporated ratio images often require multiple images as input~\cite{ShashuaRatioImage, PeersSIGGRAPH2007} or both source and target images~\cite{Stoschek2000}, limiting their real-world applicability. 
Wen~\textit{et~al.}\cite{WenRadianceMaps} propose the first work on single image face relighting with ratio images, by estimating the ratio between the radiance environment maps. 
However, they use a fixed face geometry and only consider diffuse reflections. % with manual feature correspondences
We instead estimate the ratio image between the source and target images and thus can directly produce non-diffuse relighting without resorting to a face mesh. 
Zhou~\textit{et al.}~\cite{DPR} use a ratio image based method to synthesize training data. % from CelebA-HQ~\cite{CelebA-HQ}. 
Our work is the first ratio-image based face relighting method that can model non-diffuse relighting effects including shadows caused by strong directional lights, while requiring only one source image and a target lighting as input. 

%------------------------------------------------------------------------
\Section{Proposed Method}

%------------------------------------------------------------------------
\SubSection{Architecture}

Our model takes a source image $\mathbf{I}_{s}$ and a target lighting $\mathbf{\ell}_{t}$ as input and outputs the relit image $\mathbf{I}_{r}$ along with the estimated source lighting $\mathbf{\ell}_{p}$. Our lighting representation is the first $9$ Spherical Harmonics (SH) coefficients. We adopt the hourglass network of \cite{DPR}, but rather than directly estimating the target luminance we estimate the ratio image $\mathbf{R}_{p}$ between the input and target luminance. Similar to~\cite{DPR}, our model only modifies the luminance channel: they use the \textit{Lab} color space and estimate the target luminance before recombining with the source image's \textit{a} and \textit{b} channels to generate the relit image, whereas we estimate $\mathbf{R}_{p}$ for the $Y$ channel of the \textit{YUV} color space. The $Y$ channel of the target image is computed by multiplying $\mathbf{R}_{p}$ with the $Y$ channel of the source image, which is then recombined with the \textit{U} and \textit{V}  channels of the source image and converted to \textit{RGB} to produce the final relit image (See Fig.~\ref{fig:Architecture}). 
%------------------------------------------------------------------------
\SubSection{Training Losses}

We employ several loss functions to estimate ratio images that preserve high frequency details while capturing significant changes around shadow borders. 

To directly supervise the ratio image learning, we employ the following ratio image estimation loss $L_\text{ratio}$: 
\begin{equation}
L_\text{ratio} = \frac{1}{N}\| \log_\text{10}(\mathbf{R}_{p})-\log_\text{10}(\mathbf{R}_{t}) \|_\text{1}. 
\end{equation}
Here, $\mathbf{R}_{p}$ and $\mathbf{R}_{t}$ are the predicted and ground truth ratio images respectively, and $N$ is the number of pixels in the image. Defining the loss in the $\log$ space ensures that ratio image values of $r$ and $\frac{1}{r}$ receive equal weight in the loss. 

We have two additional loss functions that place higher emphasis on the ratio image estimation near the shadow borders of the source and target images. The shadow border ratio image loss $L_\text{i,border}$ is defined as: 
\begin{equation}
L_\text{i,border} = \frac{1}{N_i}\| \mathbf{W}_{i}\odot(\log_\text{10}(\mathbf{R}_{p})-\log_\text{10}(\mathbf{R}_{t})) \|_\text{1},
\end{equation}
where $i$ is $s$ or $t$ denoting the source or target respectively, and $\odot$ is element-wise multiplication. 
$\mathbf{W}_{s}$ and $\mathbf{W}_{t}$ (See Sec.~\ref{sec:reweighting}) are per-pixel weights that are element-wise multiplied with the per-pixel ratio image error, enabling our model to emphasize the ratio image estimation at or near shadow borders. 
$N_\text{s}$ and $N_\text{t}$ are the number of pixels with nonzero weights in $\mathbf{W}_{s}$ and $\mathbf{W}_{t}$ respectively. We denote the sum of $L_\text{ratio}$, $L_\text{sborder}$, and $L_\text{tborder}$ as $L_\text{wratio}$. 

We also supervise the source lighting estimation using the loss term $L_\text{lighting}$ defined as: 
\begin{equation}
L_\text{lighting} = \| \mathbf{\ell}_{p}-\mathbf{\ell}_{s} \|^\text{2}_\text{2}, 
\end{equation}
where $\mathbf{\ell}_{p}$ and $\mathbf{\ell}_{s}$ are the predicted and ground truth source lighting, represented as the first $9$ SH coefficients. 

Similar to~\cite{DPR}, we define a gradient consistency loss $L_\text{gradient}$  to enforce that the image gradients of the predicted and target ratio images ($\mathbf{R}_{p}$ and $\mathbf{R}_{t}$) are similar, and a face feature consistency loss $L_\text{face}$ to ensure that images of the same subject under different lighting conditions have the same face features.  $L_\text{gradient}$ preserves the image edges  and avoids producing blurry relit images. $L_\text{face}$ further preserves the local facial details of the subject during relighting. 

To enhance the perceptual quality of our relit images, we employ two PatchGAN~\cite{PatchGAN} discriminators: one operates on $70\!\!\times\!\!70$ and the other on $140\!\!\times\!\!140$ patches. We train the discriminators jointly with the hourglass network using the predicted luminance as fake samples and the target luminance as real samples. We denote this loss as $L_\text{adversarial}$.

Finally, similar to \cite{PhysicsGuidedRelighting}, we define a structural dissimilarity (DSSIM) loss $L_\text{DSSIM}$ between $\mathbf{R}_{p}$ and $\mathbf{R}_{t}$ as: 
\begin{equation}
L_\text{DSSIM} = \frac{1-\text{SSIM}(\mathbf{R}_{p}, \mathbf{R}_{t})}{2}. 
\end{equation}

Our final loss function $L$ is the sum: 
\begin{align}
\begin{split}
L = L_\text{wratio}+L_\text{lighting}+L_\text{gradient}+\\
L_\text{face}+L_\text{adversarial}+L_\text{DSSIM}. 
\end{split}
\end{align} 

%------------------------------------------------------------------------
\SubSection{Shadow Border Reweighting}
\label{sec:reweighting}
We reweight our ratio image estimation loss to assign larger weights to pixels near the shadow border. 
Our reweighting scheme incorporates two key ideas. 
First, higher-contrast hard shadows should have higher weight than lower-contrast soft shadows. 
Second, the reweighted neighborhood is larger for higher contrast shadows to provide a sufficiently wide buffer zone for handling hard shadows accurately using the ratio image. Since the contrast between the shadow and illuminated regions along the shadow border is high for hard shadows, larger weights along the border are needed for our model to learn this steep gradient.

The weights should peak at the shadow border and gradually decay as we move away from the border. 
To achieve this, %generate these per-pixel weights that emphasize shadows based on our criteria, 
we first compute a shadow mask for the image using the lighting direction and $3$D shape estimated by $3$DMM fitting \cite{3DDFA,face-alignment-in-full-pose-range-a-3d-total-solution}. The shadow mask is a binary map that highlights shadow regions of the face, as detailed in Sec.~\ref{sec:shadowMask}.

We convolve the shadow mask with a $21\!\!\times\!\!21$ mean filter to get a smoothed mask $c(x,y)$. 
The smeared border can be found where the smoothed mask value is between two thresholds,  $\tau_{1} < c(x,y) < \tau_{2}$, as a border pixel's neighborhood contains a mixture of pixels inside and outside shadows. 
For each border pixel, we then compute its local contrast. Instead of using the more costly root mean squared contrast in its neighborhood, we find the sum of absolute values of the four directional derivatives along the horizontal, vertical, and two diagonal directions efficient and effective. The derivatives are obtained through $21\!\!\times\!\!21$ filters, and their sum is stored as $t(x,y)$ for each border pixel $(x,y)$. 

Given the local contrast $t(u,v)$ for each shadow border pixel $(u,v)$, 
we model its contribution to the shadow border weight at pixel $(x,y)$ as a Gaussian $g_{(u,v)}(x,y)$, which peaks at the center of the border and smoothly decays with distance away from the border: 
\begin{equation}
g_{(u,v)}(x,y) = \frac{1}{{\sigma_{max} \sqrt {2\pi } }}e^{\frac{ - \left( {c(x,y) - \mu_{c} } \right)^2 } {2\sigma_{t(u,v)} ^2 }}.
\end{equation}
Here, $\mu_{c}$ is the mean of $c(x,y)$ and $\sigma_{t(u,v)}$=$\frac{t(u,v)}{t_{max}}\sigma_{max}$ is the deviation controlling the decay speed, where $t_{max}$ is the highest contrast in the image, and $\sigma_{max}$ is the preset maximum deviation. We tally the contribution of $g_{(u,v)}$ to pixel $(x,y)$ for every border pixel $(u,v)$ whose neighborhood $N(u,v)$ contains $(x,y)$ with the neighborhood size proportional to $t(u,v)$, and obtain the shadow border weight:
\begin{equation}
\mathbf{W}(x,y) = \sum_{N(u,v)\ni (x,y)} g_{(u,v)}(x,y).
\end{equation}
As a final step, we linearly normalize $\mathbf{W}$ to the range $[0, 10]$.

\renewcommand{\thefigure}{3}
\begin{figure}[t]
%\vspace{-2mm}
\begin{center}
   \sidesubfloat[]{\includegraphics[width=0.7 \linewidth]{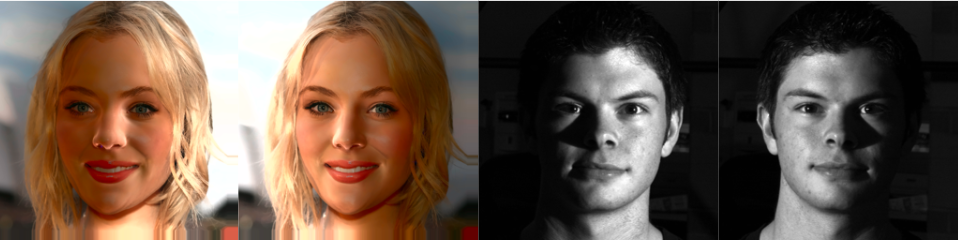}} \\
   \sidesubfloat[]{\includegraphics[width=0.7 \linewidth]{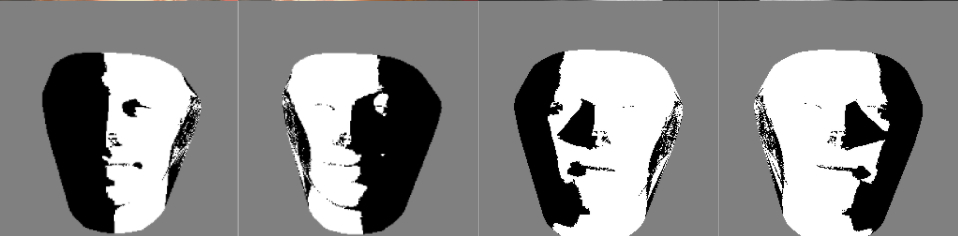}} \\
   \sidesubfloat[]{\includegraphics[width=0.7 \linewidth]{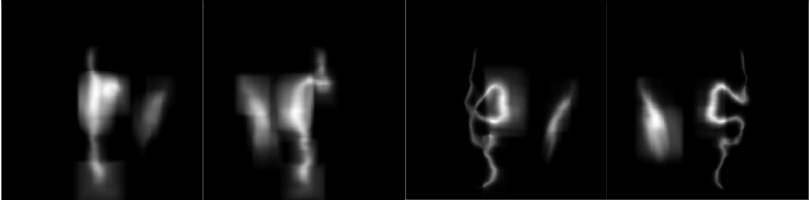}} \\
   \sidesubfloat[]{\includegraphics[width=0.7 \linewidth]{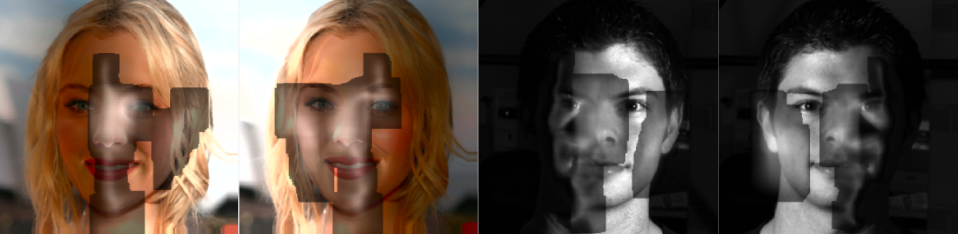}}
\vspace{3mm}
\caption{\small \textbf{Shadow Masks and Weights}. (a) input image, (b) shadow mask, (c) shadow border weights (intensity proportional to weights), (d) shadow border weights overlaid on the input image. 
For each image, we use the source lighting direction and  estimated $3$D shape to generate shadow masks indicating the shadowed regions. 
With the shadow mask, we assign high weights to learning the ratio image along the shadow border. 
Note that our reweighting scheme appropriately assigns high weights to high contrast shadows including hard cast shadows around the nose. 
}\label{fig:ShadowMask}
\end{center}\vspace{-8mm}
\end{figure}

Note that the amplitudes and standard deviations of $g_{(u,v)}$ are larger for high contrast shadow borders (hard shadows) and smaller for low contrast shadows (soft shadows). Our reweighting thus assigns the highest weights to hard shadows and their neighborhoods (See Fig.~\ref{fig:ShadowMask}) --- crucial in handling directional-light-caused shadows. % produced by strong directional lights. 

During training, we apply this reweighting scheme to both the source and target image to generate $\mathbf{W}_{s}$ and $\mathbf{W}_{t}$ respectively. This allows us to place higher emphasis on any shadow border regions in either the source or target image when predicting the ratio image. %Further details on our reweighting algorithm can be found in the supplementary materials. 

%------------------------------------------------------------------------
\SubSection{Shadow Masks}\label{sec:shadowMask}

\renewcommand{\thefigure}{4}
\begin{figure}[t]
%\vspace{-4mm}
\begin{center}
   \includegraphics[width=0.78 \linewidth]{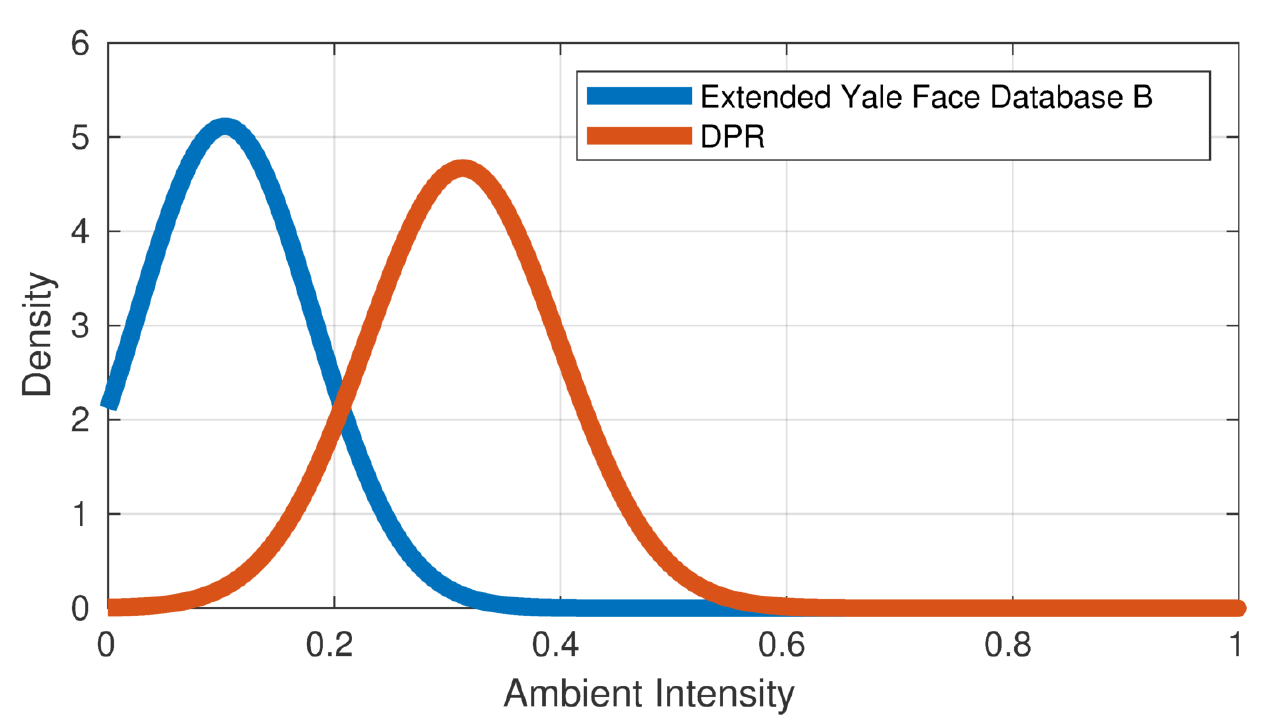}
\vspace{-2mm} \\
\caption{\small\textbf{Ambient Light Distributions.} We estimate the ambient light of our training images using shadow masks. Clearly, there is a major difference between the ambient light distributions of the images in the Extended Yale Face Database B \cite{ExtendedYale} and the images in DPR \cite{DPR}, and also a wide distribution within each dataset.  
}\label{fig:AmbientDistribution}
\vspace{-4mm}
\end{center}
\end{figure}

We create shadow masks for all training images using the lighting direction and the $3$D face mesh offered by $3$DMM fitting~\cite{3DDFA,face-alignment-in-full-pose-range-a-3d-total-solution}. We estimate the rotation, translation, and scale of the shape and align the $3$D face to the $2$D image. 

To generate the shadow mask, we cast parallel rays along the $z$-axis towards the mesh. If the ray hits the face, we check for two kinds of shadows at the intersection point: self shadows and cast shadows. Portions of the face between the light source and the intersection point will block the light and cast a shadow. To determine if the point lies in a cast shadow, we cast a ``shadow feeler'' ray from the intersection point to the light source~\cite{appel1968some}. If the ray hits the surface, the intersection point is in a cast shadow. To determine if the point lies in self shadow, we compute the angle between the light source's direction and the surface normal of the intersection point. If the angle is obtuse, the light is from the back of the surface and the point is in self shadow. If the intersection point is either in a cast shadow or self shadow, we assign $0$ to the corresponding mask pixel. Otherwise, the pixel is illuminated and is $1$ in the shadow mask. 

%------------------------------------------------------------------------
\SubSection{Estimating Ambient Light}\label{sec:Ambient}
Since ambient light is inconsistent across images, especially between light stage and in-the-wild images, we introduce a method to estimate ambient light using the shadow mask. Since only ambient light contributes to shadowed regions, we use the average intensity of the shadow pixels as an estimate of the ambient light intensity in the image. 

To sum the contributions of directional and ambient light, we first model each image's directional light as a point light and estimate the corresponding $9$ SH coefficients. We then add the estimated ambient light intensity to the $0^{\text{th}}$ order SH coefficient, which represents overall light intensity. 

During training, we use the source image's shadow mask to estimate its ambient light intensity and add that to groundtruth source lighting $\mathbf{\ell}_{s}$. We do the same using the target image's shadow mask for target lighting $\mathbf{\ell}_{t}$. 

%------------------------------------------------------------------------
\SubSection{Implementation Details}

We train our model in Tensorflow~\cite{Tensorflow} for $51$ epochs using a learning rate of $0.0001$ and the Adam optimizer~\cite{AdamOptimizer} with default parameters. Similar to~\cite{DPR}, we initially train for $8$ epochs with no skip connections, and then add them one at a time at epochs $9$, $11$, $13$, and $15$. We only apply $L_\text{face}$ after epoch $16$, whereas all other losses are always active. %During training, to effectively utilize $L_\text{face}$, we define a batch as all images for a single subject (each with a different lighting), and apply $L_\text{face}$ to all image pairs in the batch. 

Since cameras often do not correctly capture luminance in the physical world, we also apply gamma correction to the input and target luminance during training ($\gamma=\frac{1}{2.2}$) and learn the ratio image in this corrected space. 

%------------------------------------------------------------------------
\section{Experiments}

%------------------------------------------------------------------------
\subsection{Training Data}

We train our model using images from two datasets: one with mostly diffuse lighting and one with strong directional lights. Our diffuse lighting dataset is the Deep Portrait Relighting (DPR) dataset\cite{DPR}, where we use the same training images as \cite{DPR}. It contains $138,135$ images of $27,627$ subjects with $5$ lightings each. 
Our dataset with strong directional lighting is the Extended Yale Face Database B~\cite{ExtendedYale}, which contains $16,380$ images of $28$ subjects with $9$ poses and $65$ illuminations ($64$ distinct lighting directions and $1$ ambient lighting). To compensate for the large differences in ambient lighting intensities between the two datasets and within each dataset (See Fig.~\ref{fig:AmbientDistribution}), we use our method in Sec.~\ref{sec:Ambient} to produce the groundtruth source and target lightings. All quantitative and qualitative evaluations are performed using this model trained on these two databases. For training and for all evaluations, we use $256\times256$ images. 

%------------------------------------------------------------------------
\subsection{Quantitative Evaluations}

\renewcommand{\thefigure}{5}
\begin{figure*}[t]
\vspace{-2mm}
\begin{center}
\begin{minipage}[t]{0.138\linewidth}
\centering
\includegraphics[width=\linewidth]{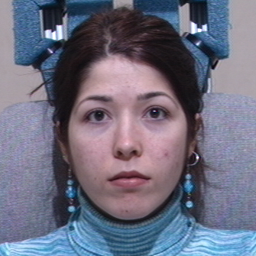}
\end{minipage}
\begin{minipage}[t]{0.138\linewidth}
\centering
\includegraphics[width=\linewidth]{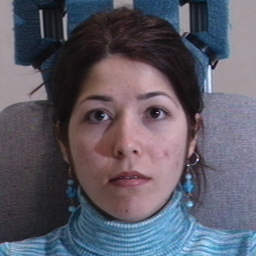}
\end{minipage}
\begin{minipage}[t]{0.138\linewidth}
\centering
\includegraphics[width=\linewidth]{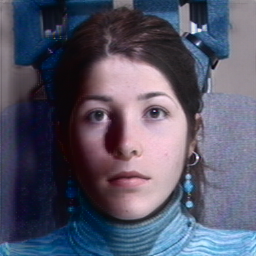}
\end{minipage}
\begin{minipage}[t]{0.138\linewidth}
\centering
\includegraphics[width=\linewidth]{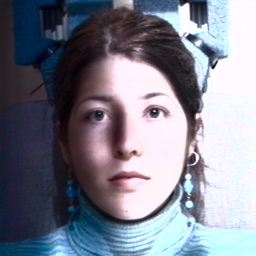}
\end{minipage}
\begin{minipage}[t]{0.138\linewidth}
\centering
\includegraphics[width=\linewidth]{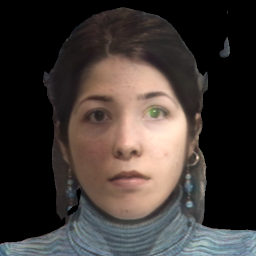}
\end{minipage}
\begin{minipage}[t]{0.138\linewidth}
\centering
\includegraphics[width=\linewidth]{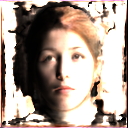}
\end{minipage}
\begin{minipage}[t]{0.138\linewidth}
\centering
\includegraphics[width=\linewidth]{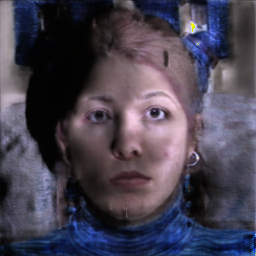}
\end{minipage}

%\newline

\begin{minipage}[t]{0.138\linewidth}
\centering
\includegraphics[width=\linewidth]{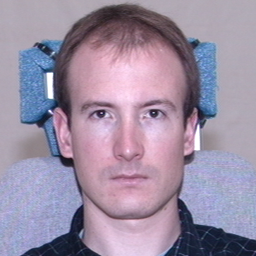} \\
\small (a) Source 
\end{minipage}
\begin{minipage}[t]{0.138\linewidth}
\centering
\includegraphics[width=\linewidth]{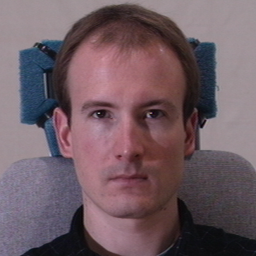} \\
\small (b) Target 
\end{minipage}
\begin{minipage}[t]{0.138\linewidth}
\centering
\includegraphics[width=\linewidth]{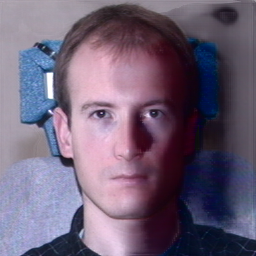} \\
\small (c) Ours
\end{minipage}
\begin{minipage}[t]{0.138\linewidth}
\centering
\includegraphics[width=\linewidth]{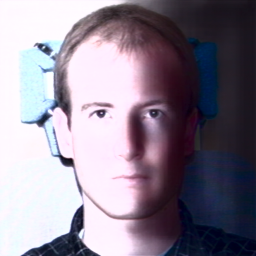} \\
\small (d) DPR \cite{DPR}
\end{minipage}
\begin{minipage}[t]{0.138\linewidth}
\centering
\includegraphics[width=\linewidth]{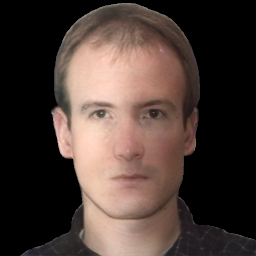} \\
\small (e) SIPR \cite{UCSDSingleImagePortraitRelighting}
\end{minipage}
\begin{minipage}[t]{0.138\linewidth}
\centering
\includegraphics[width=\linewidth]{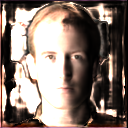} \\
\small (f) SfSNet \cite{SfSNet}
\end{minipage}
\begin{minipage}[t]{0.138\linewidth}
\centering
\includegraphics[width=\linewidth]{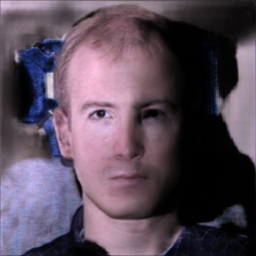} \\
\small (g) Nestmeyer \cite{PhysicsGuidedRelighting}
\end{minipage}
\vspace{-4mm}
\caption{\small \textbf{Relighting on Multi-PIE Using Target Lighting}. All methods take a source image and a target lighting as input. Images for SIPR \cite{UCSDSingleImagePortraitRelighting} are provided by the authors. Notice that our model produces significantly better cast shadows, especially around the nose. 
}\label{fig:TargetLightingMP}
\end{center}\vspace{-2mm}
\end{figure*}

\renewcommand{\thefigure}{6}
\begin{figure*}[t]
\vspace{-2mm}
\begin{center}
\begin{minipage}[t]{0.12\linewidth}
\centering
\includegraphics[width=\linewidth]{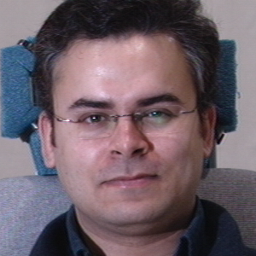}
\end{minipage}
\begin{minipage}[t]{0.12\linewidth}
\centering
\includegraphics[width=\linewidth]{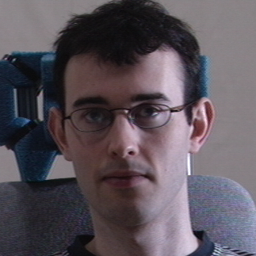}
\end{minipage}
\begin{minipage}[t]{0.12\linewidth}
\centering
\includegraphics[width=\linewidth]{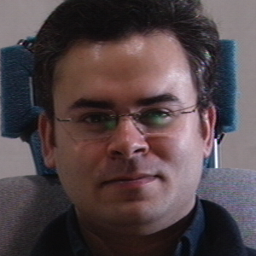}
\end{minipage}
\begin{minipage}[t]{0.12\linewidth}
\centering
\includegraphics[width=\linewidth]{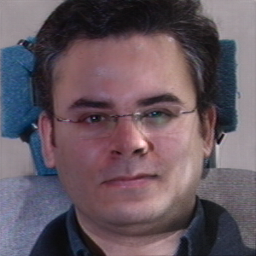}
\end{minipage}
\begin{minipage}[t]{0.12\linewidth}
\centering
\includegraphics[width=\linewidth]{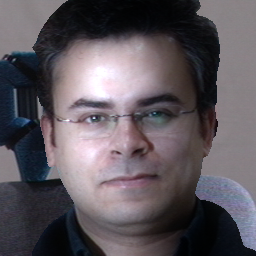}
\end{minipage}
\begin{minipage}[t]{0.12\linewidth}
\centering
\includegraphics[width=\linewidth]{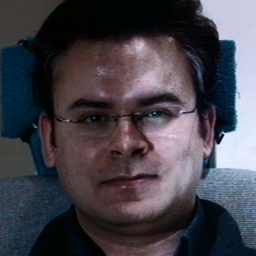}
\end{minipage}

%\newline 

\begin{minipage}[t]{0.12\linewidth}
\centering
\includegraphics[width=\linewidth]{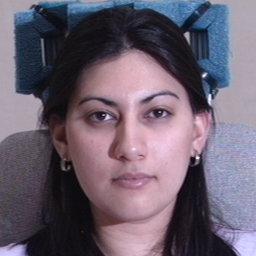} \\
\small (a) Source 
\end{minipage}
\begin{minipage}[t]{0.12\linewidth}
\centering
\includegraphics[width=\linewidth]{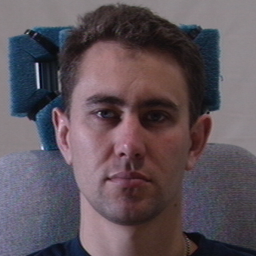} \\
\small (b) Reference
\end{minipage}
\begin{minipage}[t]{0.12\linewidth}
\centering
\includegraphics[width=\linewidth]{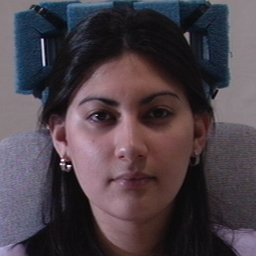} \\
\small (c) Target
\end{minipage}
\begin{minipage}[t]{0.12\linewidth}
\centering
\includegraphics[width=\linewidth]{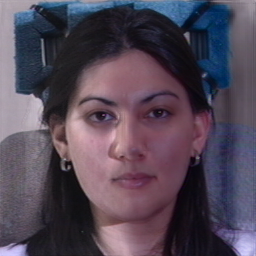} \\
\small (d) Ours 
\end{minipage}
\begin{minipage}[t]{0.12\linewidth}
\centering
\includegraphics[width=\linewidth]{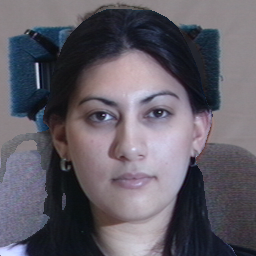} \\
\small (e) Shih \cite{Flickr}
\end{minipage}
\begin{minipage}[t]{0.12\linewidth}
\centering
\includegraphics[width=\linewidth]{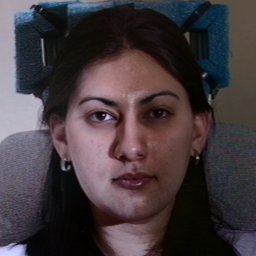} \\
\small (f) Shu \cite{MassTransport}
\end{minipage}
\vspace{-1mm}
\caption{\small \textbf{Lighting Transfer on Multi-PIE}. All methods estimate the target lighting from the reference. Images of~\cite{MassTransport} are provided by the authors. Our model produces the correct lightings, whereas \cite{Flickr} does not. Unlike ours, both baselines strongly alter the skin tones. %, whereas ours does not. 
}\label{fig:LightingTransfer}
\end{center}\vspace{-5mm}
\end{figure*}

We compare our model's relighting performance with prior work on the Multi-PIE dataset \cite{Multi-PIE}, where each subject is illuminated under $19$ lighting conditions ($18$ images with directional lights, one image with only ambient lighting). 

\renewcommand{\thefigure}{7}
\begin{figure*}[t]
\vspace{-2mm}
\begin{center}
\begin{minipage}[t]{0.11\linewidth}
\centering
\includegraphics[width=\linewidth]{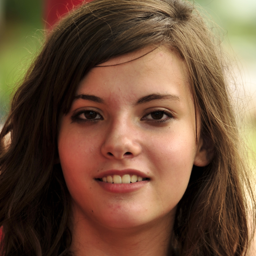}
\end{minipage}
\begin{minipage}[t]{0.11\linewidth}
\centering
\includegraphics[width=\linewidth]{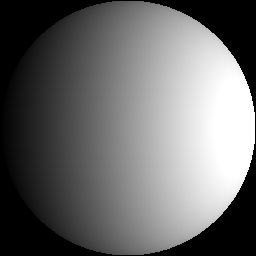}
\end{minipage}
\begin{minipage}[t]{0.11\linewidth}
\centering
\includegraphics[width=\linewidth]{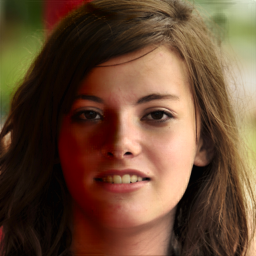}
\end{minipage}
\begin{minipage}[t]{0.11\linewidth}
\centering
\includegraphics[width=\linewidth]{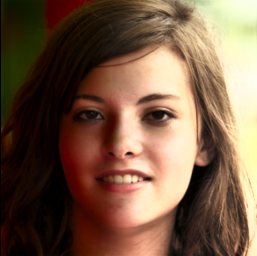}
\end{minipage}
\begin{minipage}[t]{0.11\linewidth}
\centering
\includegraphics[width=\linewidth]{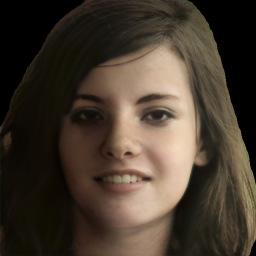}
\end{minipage}
\begin{minipage}[t]{0.11\linewidth}
\centering
\includegraphics[width=\linewidth]{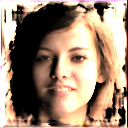}
\end{minipage}

%\newline 

\begin{minipage}[t]{0.11\linewidth}
\centering
\includegraphics[width=\linewidth]{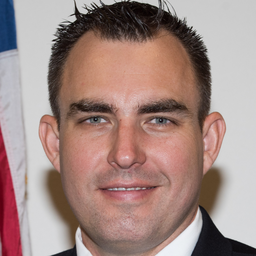}
\end{minipage}
\begin{minipage}[t]{0.11\linewidth}
\centering
\includegraphics[width=\linewidth]{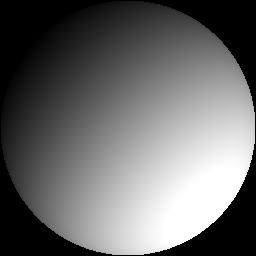}
\end{minipage}
\begin{minipage}[t]{0.11\linewidth}
\centering
\includegraphics[width=\linewidth]{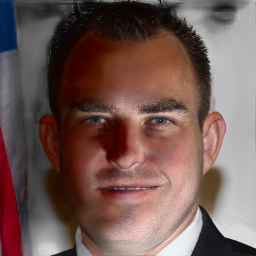}
\end{minipage}
\begin{minipage}[t]{0.11\linewidth}
\centering
\includegraphics[width=\linewidth]{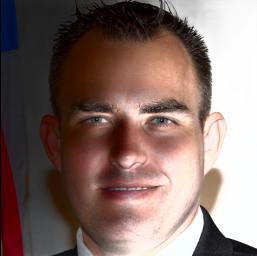}
\end{minipage}
\begin{minipage}[t]{0.11\linewidth}
\centering
\includegraphics[width=\linewidth]{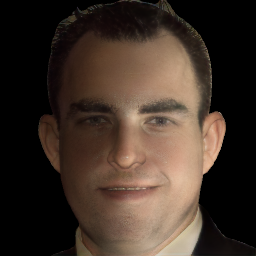}
\end{minipage}
\begin{minipage}[t]{0.11\linewidth}
\centering
\includegraphics[width=\linewidth]{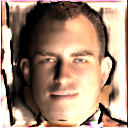}
\end{minipage}

%\newline 

\begin{minipage}[t]{0.11\linewidth}
\centering
\includegraphics[width=\linewidth]{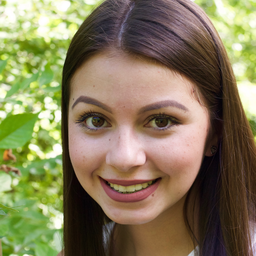}
\end{minipage}
\begin{minipage}[t]{0.11\linewidth}
\centering
\includegraphics[width=\linewidth]{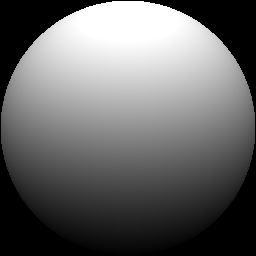}
\end{minipage}
\begin{minipage}[t]{0.11\linewidth}
\centering
\includegraphics[width=\linewidth]{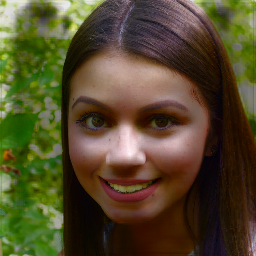}
\end{minipage}
\begin{minipage}[t]{0.11\linewidth}
\centering
\includegraphics[width=\linewidth]{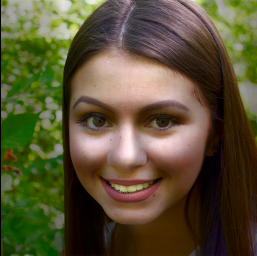}
\end{minipage}
\begin{minipage}[t]{0.11\linewidth}
\centering
\includegraphics[width=\linewidth]{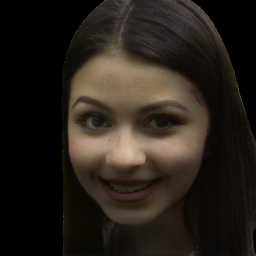}
\end{minipage}
\begin{minipage}[t]{0.11\linewidth}
\centering
\includegraphics[width=\linewidth]{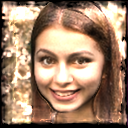}
\end{minipage}

%\newline 

\begin{minipage}[t]{0.11\linewidth}
\centering
\includegraphics[width=\linewidth]{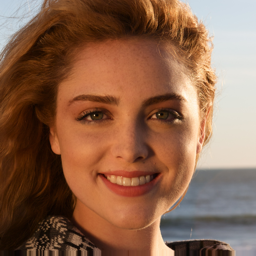}
\end{minipage}
\begin{minipage}[t]{0.11\linewidth}
\centering
\includegraphics[width=\linewidth]{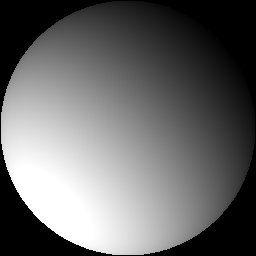}
\end{minipage}
\begin{minipage}[t]{0.11\linewidth}
\centering
\includegraphics[width=\linewidth]{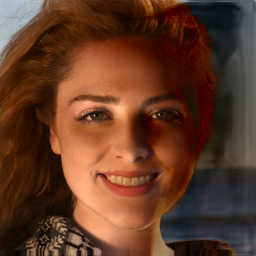}
\end{minipage}
\begin{minipage}[t]{0.11\linewidth}
\centering
\includegraphics[width=\linewidth]{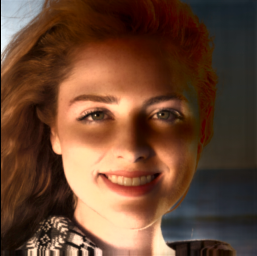}
\end{minipage}
\begin{minipage}[t]{0.11\linewidth}
\centering
\includegraphics[width=\linewidth]{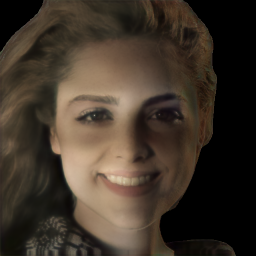}
\end{minipage}
\begin{minipage}[t]{0.11\linewidth}
\centering
\includegraphics[width=\linewidth]{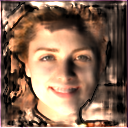}
\end{minipage}

%\newline 

\begin{minipage}[t]{0.11\linewidth}
\centering
\includegraphics[width=\linewidth]{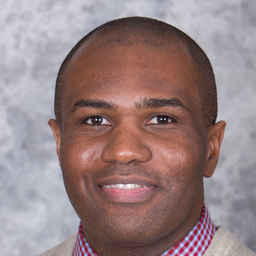} \\
\small (a) Source 
\end{minipage}
\begin{minipage}[t]{0.11\linewidth}
\centering
\includegraphics[width=\linewidth]{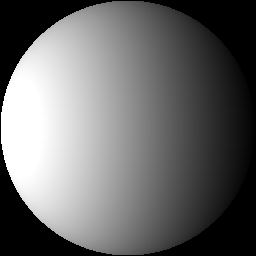} \\
\small (b) Target
\end{minipage}
\begin{minipage}[t]{0.11\linewidth}
\centering
\includegraphics[width=\linewidth]{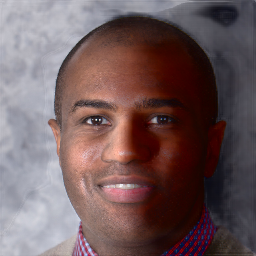} \\
\small (c) Our model
\end{minipage}
\begin{minipage}[t]{0.11\linewidth}
\centering
\includegraphics[width=\linewidth]{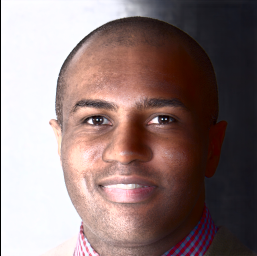} \\
\small (d) DPR \cite{DPR} 
\end{minipage}
\begin{minipage}[t]{0.11\linewidth}
\centering
\includegraphics[width=\linewidth]{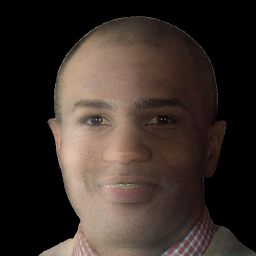} \\
\small (e) SIPR \cite{UCSDSingleImagePortraitRelighting}
\end{minipage}
\begin{minipage}[t]{0.11\linewidth}
\centering
\includegraphics[width=\linewidth]{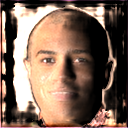} \\
\small (f) SfSNet \cite{SfSNet} 
\end{minipage}
\caption{\small\textbf{Relighting on FFHQ dataset Using Target Lighting}. %We compare our relighting results on FFHQ subjects with prior work. 
Images for SIPR \cite{UCSDSingleImagePortraitRelighting} are provided by the authors. Across all lighting conditions, our model produces better cast shadows than prior work, especially around the nose and eyes.  
}\label{fig:FFHQQualitative}
\end{center}\vspace{-6mm}
\end{figure*}

%------------------------------------------------------------------------
\noindent \textbf{Relighting Using Target Lightings.}  We compare against prior relighting methods \cite{DPR, SfSNet, PhysicsGuidedRelighting} that can accept a target lighting as input. For each Multi-PIE subject and each session, we randomly select one of the $19$ images as the source image and one as the target image, which serves as the relighting groundtruth. The target image's lighting is then used to relight the source image. This leads to a total of $921$ relit images. For prior methods, we use the pretrained models provided by the authors. We evaluate the performance using $3$ error metrics: Si-MSE \cite{DPR}, MSE, and DSSIM, as shown in Tab.~\ref{tab:TargetLighting}. Unlike traditional MSE, Si-MSE computes the MSE between the target image and the relit image multiplied by the scale that yields the minimum error for the image. This is meant to resolve scale ambiguities in the relit image caused by the target lighting. We don't compare quantitatively with SIPR \cite{UCSDSingleImagePortraitRelighting} since their method masks out the background. Most notably, our model outperforms DPR~\cite{DPR} while using their model architecture, showing the benefits of the ratio image and our shadow border losses. 

\begin{table}[t!]

\begin{center}
\tiny
\scalebox{1.23}{
\setlength\tabcolsep{1.5pt}{  
\begin{tabular}{| c | c | c | c |}
\hline
Method & Si-MSE & MSE & DSSIM \\
\hline
SfSNet \cite{SfSNet} & $0.0545\!\pm\!0.0201$ &$0.1330\!\pm\!0.0531$ & $0.3151\!\pm\!0.0523$ \\
\hline
DPR \cite{DPR} & $0.0282\!\pm\!0.0125$ & $0.0702\!\pm\!0.0361$ & $0.1818\!\pm\!0.0490$\\
\hline
Nestmeyer \cite{PhysicsGuidedRelighting} & $0.0484\!\pm\!0.0318$ & $0.0584\!\pm\!0.0335$ & $0.2722\!\pm\!0.0950$ \\
\hline
Our model & $\mathbf{0.0220\!\pm\!0.0078}$ & $\mathbf{0.0292\!\pm\!0.0148}$ & $\mathbf{0.1605\!\pm\!0.0487}$ \\
\hline
\end{tabular}
}}
\end{center}

\vspace{-3mm}
\caption{\small
\textbf{Relighting Results (mean$\pm$ standard deviation) Using Target Lighting on Multi-PIE}. All methods take a source image and a target lighting as input. Our method outperforms prior works across all metrics, and also has the lowest standard deviation.   
}\label{tab:TargetLighting}
\vspace{-2mm}
\end{table}
%------------------------------------------------------------------------
\noindent \textbf{Relighting Using Reference Images.} We also compare with relighting methods that require both a source and a reference image as input \cite{Flickr, MassTransport}, which relight by transferring the reference's lighting to the source. For each Multi-PIE image, we randomly select a reference image across all subjects of the dataset and estimate the target lighting from the reference. %We then relight the source using the estimated target lighting. 
We use the available code to compare against Shih \textit{et al.} \cite{Flickr} and the results for Shu \textit{et al.} \cite{MassTransport} are provided by the authors. The evaluation is shown in Tab.~\ref{tab:LightingTransfer}. 

%------------------------------------------------------------------------
\noindent \textbf{Facial Detail Preservation.} To compare our model's ability to preserve the subject's facial details during relighting with prior work, we compute the average cosine similarity of the VGG-Face \cite{VGGFace} features of the relit and groundtruth Multi-PIE images across different layers of VGG-Face. In particular, we use the layer before each max-pooling layer in VGG-Face. As shown in Fig.~\ref{fig:VGGFaceFeatureSimilarity}, our model achieves noticeably higher cosine similarity than prior work in earlier layers and is comparable or slightly better in later layers, indicating that ours is better at preserving local facial details. 

\begin{table}[t!]
\begin{center}
\tiny
\scalebox{1.31}{
\setlength\tabcolsep{1.5pt}{  
\begin{tabular}{| c | c | c | c |}
\hline
Method & Si-MSE & MSE & DSSIM \\
\hline
Shih \cite{Flickr} & $0.0374\!\pm\!0.0190$ & $0.0455\!\pm\!0.0203$ & $0.2260\!\pm\!0.0443$ \\
\hline
Shu \cite{MassTransport} & $0.0162\!\pm\!0.0102$ & $0.0243\!\pm\!0.0170$ & $0.1383\!\pm\!0.0499$ \\
\hline
Our model & $\mathbf{0.0148\!\pm\!0.0096}$ & $\mathbf{0.0204\!\pm\!0.0153}$ & $\mathbf{0.1150\!\pm\!0.0404}$ \\
\hline
\end{tabular}
}}
\end{center}
\vspace{-5mm}
\caption{\small
\textbf{Lighting Transfer Results (mean$\pm$ standard deviation) on Multi-PIE}.  Each input image is assigned a random reference image. Our model outperforms both methods in all metrics.  
}
\label{tab:LightingTransfer}
%\vspace{2mm}
\end{table}

\renewcommand{\thefigure}{8}
\begin{figure}[t]
\vspace{-2mm}
\begin{center}
   \includegraphics[width=0.82 \linewidth]{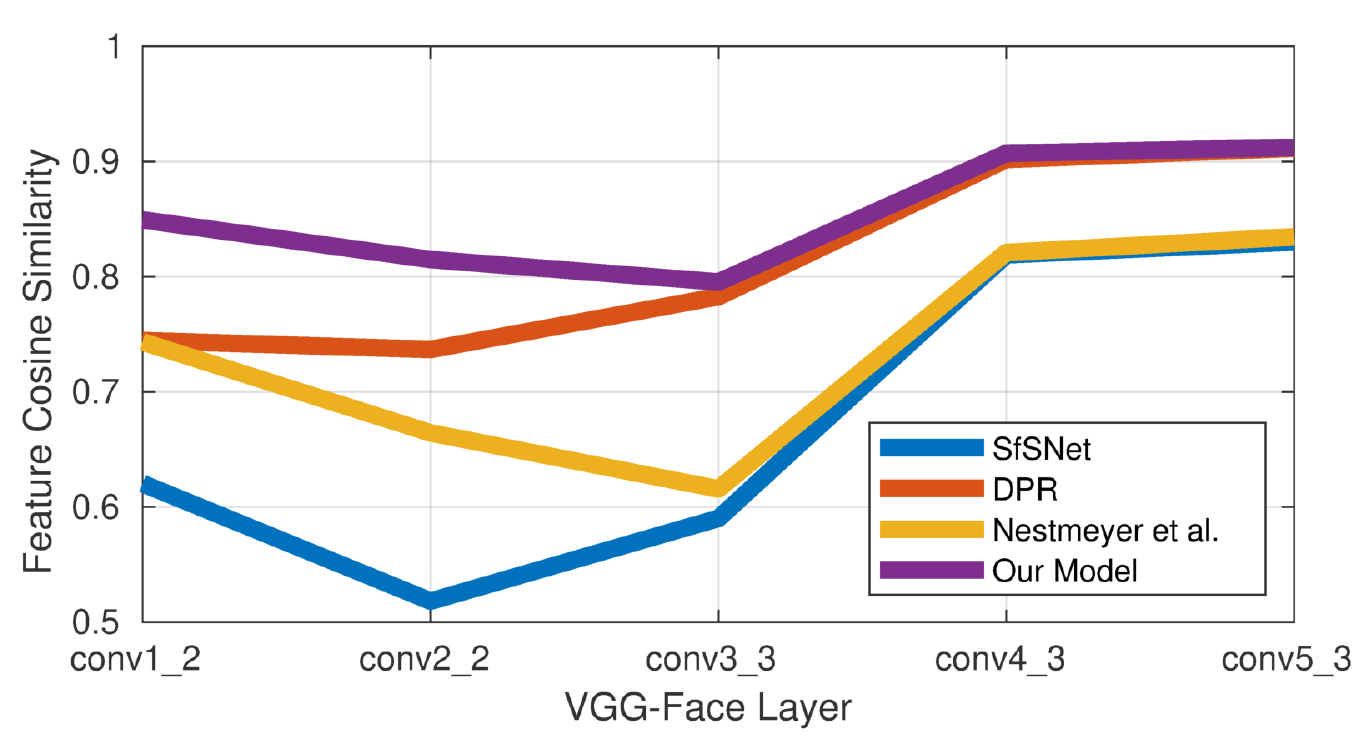}
\vspace{-6mm}
\caption{\small \textbf{Facial Detail Preservation on Multi-PIE}. We compute the cosine similarity between the VGG-Face~\cite{VGGFace} features of each method's relit images and groundtruth target images to measure their ability to preserve local and global facial details during relighting. Our method is consistently the best at preserving local details based on the cosine similarity in the {\it earlier} layers. 
}\label{fig:VGGFaceFeatureSimilarity}
\end{center} %\vspace{-4mm}
\end{figure}

%------------------------------------------------------------------------
\vspace{-1mm}
\subsection{Qualitative Evaluations}

We qualitatively compare relighting results on Multi-PIE~\cite{Multi-PIE} and FFHQ~\cite{FFHQ}. % and compare with prior work. 
On Multi-PIE, we include relit images produced from target lightings (See Fig.~\ref{fig:TargetLightingMP}) and from lighting transfer (See Fig.~\ref{fig:LightingTransfer}). When applying target lighting, our model is able to produce noticeably better cast shadows than DPR \cite{DPR}, SIPR \cite{UCSDSingleImagePortraitRelighting}, SfSNet \cite{SfSNet}, and Nestmeyer \textit{et al.} \cite{PhysicsGuidedRelighting}. Our model also avoids overlighting the image, whereas \cite{DPR} often produces images that appear overexposed. When performing lighting transfer, Shih \textit{et al.} \cite{Flickr} is unable to produce the correct lighting whereas ours can transfer the correct target lighting. Both baselines also greatly alter the subjects' skin tones, whereas our model does not. 

On FFHQ, we perform relighting using target lighting (See Fig.~\ref{fig:FFHQQualitative}). Our approach handles cast shadows better than prior work, as seen by the shadows around the nose, eyes, and other regions, while also maintaining similar or better visual quality in the relit images. When evaluating Nestmeyer \textit{et al.} \cite{PhysicsGuidedRelighting}, we find that it does not generalize well to the in-the-wild images in FFHQ due to their training dataset and thus we do not include it in Fig.~\ref{fig:FFHQQualitative}. We discuss this further in the supplementary materials. 

%------------------------------------------------------------------------
\subsection{Ablation Studies}

\renewcommand{\thefigure}{9}
\begin{figure}[t]
\vspace{-2mm}
\begin{center}
\begin{minipage}[t]{0.24\linewidth}
\centering
\includegraphics[width=\linewidth]{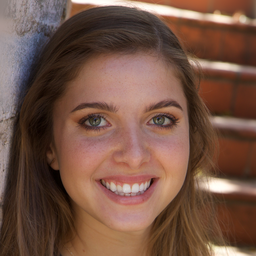}
\end{minipage}
\begin{minipage}[t]{0.24\linewidth}
\centering
\includegraphics[width=\linewidth]{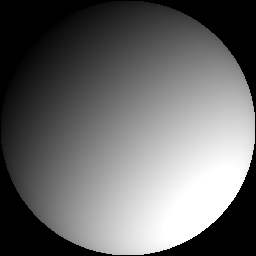}
\end{minipage}
\begin{minipage}[t]{0.24\linewidth}
\centering
\includegraphics[width=\linewidth]{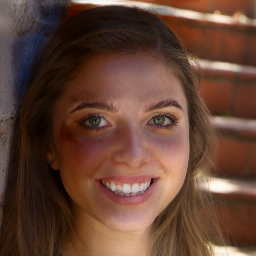}
\end{minipage}
\begin{minipage}[t]{0.24\linewidth}
\centering
\includegraphics[width=\linewidth]{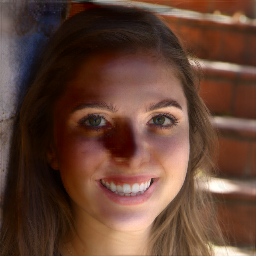}
\end{minipage}

%\newline 

\begin{minipage}[t]{0.24\linewidth}
\centering
\includegraphics[width=\linewidth]{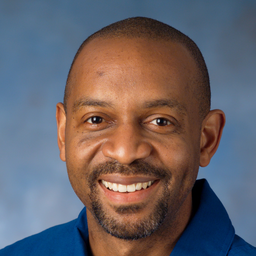}
\small (a) Input
\end{minipage}
\begin{minipage}[t]{0.24\linewidth}
\centering
\includegraphics[width=\linewidth]{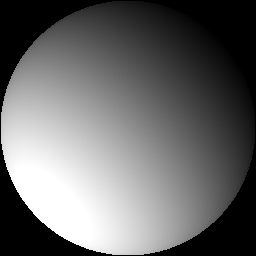}
\small (b) Target
\end{minipage}
\begin{minipage}[t]{0.24\linewidth}
\centering
\includegraphics[width=\linewidth]{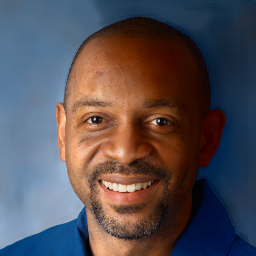}
\small (c) No $L_\text{s/tborder}$
\end{minipage}
\begin{minipage}[t]{0.24\linewidth}
\centering
\includegraphics[width=\linewidth]{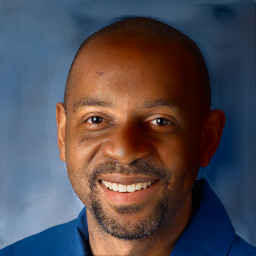}
\small (d) Proposed
\end{minipage}
\vspace{2mm}
\caption{\small \textbf{Shadow Border Ablation}. The shadow border losses improve the quality of cast shadows, especially around the nose. 
}\label{fig:ShadowBorderAblation}
\end{center}\vspace{-8mm}
\end{figure}

\begin{table}[t!]
\begin{center}
%\small
\scalebox{0.85}{
\setlength\tabcolsep{5.25pt}{  
\begin{tabular}{| c | c | c | c |}
\hline
Method & Si-MSE & MSE & DSSIM \\
\hline
No shadow border & $0.0212$ & $0.0290$ & $0.1528$ \\
\hline
No ratio image & $0.0230$ & $0.0343$ & $0.1617$ \\
\hline
Proposed & $0.0220$ & $0.0292$ & $0.1605$ \\
\hline
\end{tabular}
}}
\end{center}
\vspace{-8mm}
 \caption{\small
\textbf{Ablation on Multi-PIE}. We ablate on our Multi-PIE test set against our model with no shadow border losses and our model that estimates the luminance directly rather than the ratio image.   
}\label{tab:Ablations}
\vspace{2mm}
\end{table}

%------------------------------------------------------------------------
\noindent \textbf{Shadow Border Reweighting.} To demonstrate the effectiveness of our reweighting of the ratio image estimation loss, we train an additional model without $L_\text{sborder}$ and $L_\text{tborder}$. Our proposed model is able to produce significantly better cast shadows, especially around the nose region, whereas the model trained without $L_\text{sborder}$ and $L_\text{tborder}$ often cannot produce cast shadows (see Fig.~\ref{fig:ShadowBorderAblation}). This highlights the contribution of our shadow border losses. We also quantitatively evaluate both models on Multi-PIE, as shown in Tab.~\ref{tab:Ablations}. The model with no shadow border weights appears to slightly outperform the proposed model quantitatively, however we've shown in Fig.~\ref{fig:ShadowBorderAblation} that it cannot produce cast shadows. 
The small difference is likely due to the emphasis that the border weights place on the shadow borders over other pixels in the image, resulting in a tradeoff between high quality cast shadows and overall visual quality. 

%------------------------------------------------------------------------
\noindent \textbf{Ratio Image Estimation.} We perform a second ablation to show the benefits of modeling the relighting problem as a ratio image estimation problem. We train a second model that estimates the luminance directly rather than the ratio image. As shown in Fig.~\ref{fig:RatioImageAblation}, our proposed model is able to better preserve the local facial details of the subject in several regions, particularly around the mouth. We also quantitatively evaluate both models on Multi-PIE, as shown in Tab.~\ref{tab:Ablations}. Our proposed model performs better across all metrics. This validates that estimating the ratio image rather than the luminance improves the fidelity of the relit images. \\
%------------------------------------------------------------------------
\noindent \textbf{Ambient Lighting Estimation.} We ablate the benefits of modeling the different ambient lighting intensities across different datasets, which enables our model to train with multiple datasets. We train an additional model that does not account for the ambient lighting differences between our two datasets and instead only uses the SH lighting estimate from the point light source. Our proposed model produces more accurate global lighting intensities on Multi-PIE, especially around the face, showing the benefit of our ambient lighting estimation (See Fig.~\ref{fig:AmbientAblation}). To demonstrate the performance quantitatively, we compute the MSE in the face region using portrait masks. Our proposed model achieves a lower MSE of $0.0233$ compared to our model trained with no ambient lighting estimation, which achieves $0.0242$. 
%\vspace{-5mm}
%------------------------------------------------------------------------
\section{Conclusion}

\renewcommand{\thefigure}{10}
\begin{figure}[t]
\vspace{-2mm}
\begin{center}
\begin{minipage}[t]{0.24\linewidth}
\centering
\includegraphics[width=\linewidth]{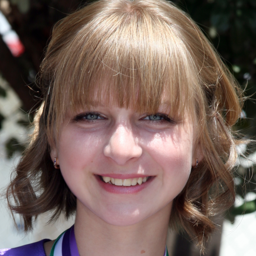}
\end{minipage}
\begin{minipage}[t]{0.24\linewidth}
\centering
\includegraphics[width=\linewidth]{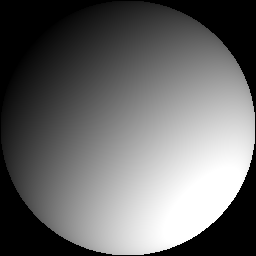}
\end{minipage}
\begin{minipage}[t]{0.24\linewidth}
\centering
\includegraphics[width=\linewidth]{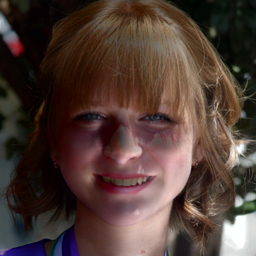}
\end{minipage}
\begin{minipage}[t]{0.24\linewidth}
\centering
\includegraphics[width=\linewidth]{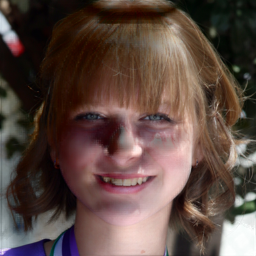}
\end{minipage}

%\newline 

\begin{minipage}[t]{0.24\linewidth}
\centering
\includegraphics[width=\linewidth]{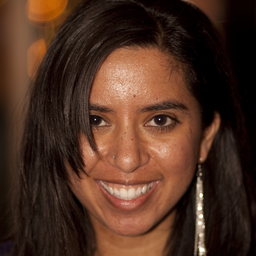}
\small (a) Input
\end{minipage}
\begin{minipage}[t]{0.24\linewidth}
\centering
\includegraphics[width=\linewidth]{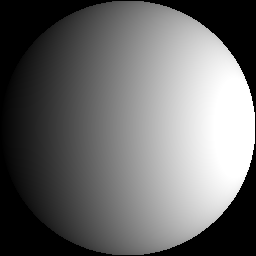}
\small (b) Target
\end{minipage}
\begin{minipage}[t]{0.24\linewidth}
\centering
\includegraphics[width=\linewidth]{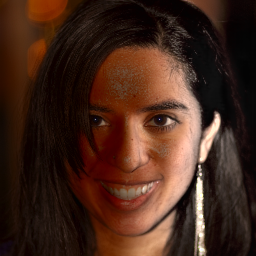}
\small (c) No Ratio
\end{minipage}
\begin{minipage}[t]{0.24\linewidth}
\centering
\includegraphics[width=\linewidth]{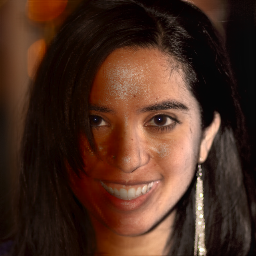}
\small (d) Proposed
\end{minipage}
\vspace{-2mm}
\caption{\small \textbf{Ratio Image Ablation}. d) better preserves the local facial details of the subject, most notably around the mouth. In the case of the bottom subject, c) seems to be gazing to the left instead of forward which is fixed by d). Best viewed if enlarged. 
}\label{fig:RatioImageAblation}
\end{center}\vspace{-4mm}
\end{figure}

\renewcommand{\thefigure}{11}
\begin{figure}[t]
\vspace{-2mm}
\begin{center}
\begin{minipage}[t]{0.24\linewidth}
\centering
\includegraphics[width=\linewidth]{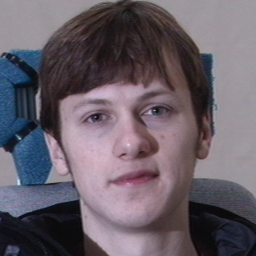}
\end{minipage}
\begin{minipage}[t]{0.24\linewidth}
\centering
\includegraphics[width=\linewidth]{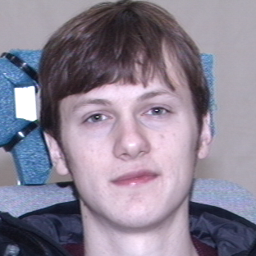}
\end{minipage}
\begin{minipage}[t]{0.24\linewidth}
\centering
\includegraphics[width=\linewidth]{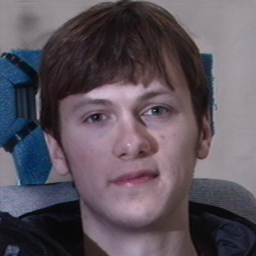}
\end{minipage}
\begin{minipage}[t]{0.24\linewidth}
\centering
\includegraphics[width=\linewidth]{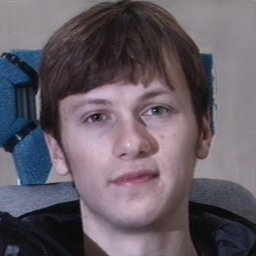}
\end{minipage}

%\newline 

\begin{minipage}[t]{0.24\linewidth}
\centering
\includegraphics[width=\linewidth]{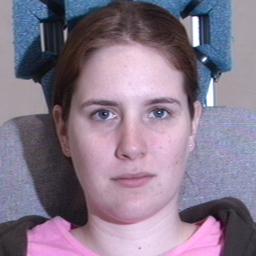}
\small (a) Input
\end{minipage}
\begin{minipage}[t]{0.24\linewidth}
\centering
\includegraphics[width=\linewidth]{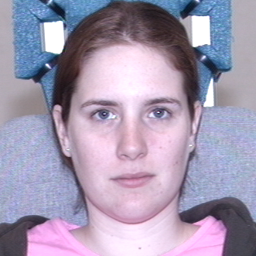}
\small (b) Target
\end{minipage}
\begin{minipage}[t]{0.24\linewidth}
\centering
\includegraphics[width=\linewidth]{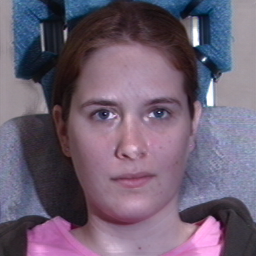}
\small (c) No Ambient
\end{minipage}
\begin{minipage}[t]{0.24\linewidth}
\centering
\includegraphics[width=\linewidth]{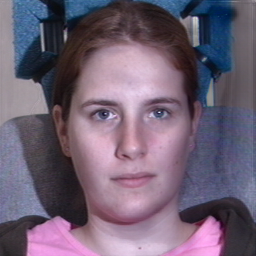}
\small (d) Proposed
\end{minipage}
\vspace{2mm}
\caption{\small \textbf{Ambient Light Ablation}. d) produces a more accurate global lighting intensity than c), especially around the face, which shows the benefit of our ambient light estimation method. 
}\label{fig:AmbientAblation}
\end{center}\vspace{-8mm}
\end{figure}

We have proposed a novel face relighting method that creates realistic hard shadows. We introduce the idea of improving shadow modeling by increasing the emphasis on correctly relighting the pixels near shadow borders, and provide an algorithm to locate the borders and reweight losses around them using shadow masks. We also introduce the first ratio-image based face relighting method that both models the effects of strong directional lights and requires only a single source image and a target lighting as input, and we demonstrate our model's ability to better preserve the local facial details of the subject as a result of using the ratio image. We further introduce a method to estimate the ambient lighting intensity using an image's shadow mask, allowing our model to account for differences in environmental lighting across datasets. Our model can thus benefit from multiple datasets during training, which improves its generalizability to unseen data. We show that our shadow border reweighting scheme, our use of the ratio image, and our ambient light estimation all improve the relighting performance. We hope that this work will inspire future relighting methods, especially with respect to shadow handling. 

\vspace{2mm}
\noindent\textbf{Acknowledgement}
This work was partially funded by Qualcomm Technologies Inc.
The authors thank Dr.~Zhixin Shu and Tiancheng Sun for kindly providing the results for their methods.

{\small
\bibliographystyle{ieee_fullname}
\bibliography{egbib}

\begin{thebibliography}{10}\itemsep=-1pt

\bibitem{Tensorflow}
Martin Abadi, Paul Barham, Jianmin Chen, Zhifeng Chen, Andy Davis, Jeffrey
  Dean, et~al.
\newblock Tensorflow: A system for large-scale machine learning.
\newblock In {\em OSDI}, 2016.

\bibitem{appel1968some}
Arthur Appel.
\newblock Some techniques for shading machine renderings of solids.
\newblock In {\em AFIPS}, 1968.

\bibitem{BarronMalikSfS}
Jonathan~T Barron and Jitendra Malik.
\newblock Shape, illumination, and reflectance from shading.
\newblock {\em PAMI}, 2015.

\bibitem{3DMM}
Volker Blanz and Thomas Vetter.
\newblock A morphable model for the synthesis of 3{D} faces.
\newblock In {\em SIGGRAPH}, 1999.

\bibitem{EggerIJCV2018}
Bernhard Egger, Sandro Schonborn, Andreas Schneider, Adam Kortylewski, Andreas
  Morel-Forseter, Clemens Blumer, and Thomas Vetter.
\newblock Occlusion-aware 3{D} morphable models and an illumination prior for
  face image analysis.
\newblock {\em IJCV}, 2018.

\bibitem{FreemanIntrinsic2018}
Kyle Genova, Forrester Cole, Aaron Maschinot, Aaron Sarna, Daniel Vlasic, and
  William~T. Freeman.
\newblock Unsupervised training for 3{D} morphable model regression.
\newblock In {\em CVPR}, 2018.

\bibitem{ExtendedYale}
Athinodoros Georghiades, Peter Belhumeur, and David Kriegman.
\newblock From {F}ew to {M}any: {I}llumination {C}one {M}odels for {F}ace
  {R}ecognition under {V}ariable {L}ighting and {P}ose.
\newblock {\em PAMI}, 2001.

\bibitem{PatchGAN}
Ian Goodfellow, Jean Pouget-Abadie, Mehdi Mirza, Bing Xu, David Warde-Farley,
  Sherjil Ozair, Aaron Courville, and Yoshua Bengio.
\newblock Generative adversarial nets.
\newblock In {\em NeurIPS}, 2014.

\bibitem{Multi-PIE}
Ralph Gross, Iain Matthews, Jeffrey Cohn, Takeo Kanade, and Simon Baker.
\newblock Multi-{PIE}.
\newblock {\em Image and Vision Computing}, 2010.

\bibitem{improving-face-recognition-from-hard-samples-via-distribution-distillation-loss}
Yuge Huang, Pengcheng Shen, Ying Tai, Shaoxin Li, Xiaoming Liu, Jilin Li,
  Feiyue Huang, and Rongrong Ji.
\newblock Improving face recognition from hard samples via distribution
  distillation loss.
\newblock In {\em ECCV}, 2020.

\bibitem{FFHQ}
Tero Karras, Samuli Laine, and Timo Aila.
\newblock A style-based generator architecture for generative adversarial
  networks.
\newblock In {\em CVPR}, 2019.

\bibitem{AdamOptimizer}
Diederik Kingma and Jimmy~Ba. Adam.
\newblock A method for stochastic optimization.
\newblock In {\em ICLR}, 2014.

\bibitem{HaLeWACV19}
Ha Le and Ioannis Kakadiaris.
\newblock Illumination-invariant face recognition with deep relit face images.
\newblock In {\em WACV}, 2019.

\bibitem{UncertaintyFaceReconstruction}
Gun-Hee Lee and Seong-Whan Lee.
\newblock Uncertainty-{A}ware {M}esh {D}ecoder for {H}igh {F}idelity 3{D}
  {F}ace {R}econstruction.
\newblock In {\em CVPR}, 2020.

\bibitem{Pfister}
Jinho Lee, Raghu Machiraju, Baback Moghaddam, and Hanspeter Pfister.
\newblock Estimation of 3{D} faces and illumination from single photographs
  using a bilinear illumination model.
\newblock In {\em EGSR}, 2005.

\bibitem{IntrinsicFacePriors}
Chen Li, Kun Zhou, and Stephen Lin.
\newblock Intrinsic face image decomposition with human face priors.
\newblock In {\em ECCV}, 2014.

\bibitem{ClosedFormSolution}
Yijun Li, Ming-Yu Liu, Xueting Li, Ming-Hsuan Yang, and Jan Kautz.
\newblock A closed-form solution to photorealistic image stylization.
\newblock In {\em ECCV}, 2018.

\bibitem{TowardsHighFidelityFaceReconstruction}
Jiangke Lin, Yi Yuan, Tianjia Shao, and Kun Zhou.
\newblock Towards {H}igh-{F}idelity 3{D} {F}ace {R}econstruction from
  {I}n-the-{W}ild {I}mages {U}sing {G}raph {C}onvolutional {N}etworks.
\newblock In {\em CVPR}, 2020.

\bibitem{DeepPhotoStyleTransfer}
Fujun Luan, Sylvain Paris, Eli Shechtman, and Kavita Bala.
\newblock Deep photo style transfer.
\newblock In {\em CVPR}, 2017.

\bibitem{PhysicsGuidedRelighting}
Thomas Nestmeyer, Jean-Francois Lalonde, Iain Matthews, and Andreas Lehrmann.
\newblock Learning {P}hysics-guided {F}ace {R}elighting under {D}irectional
  {L}ight.
\newblock In {\em CVPR}, 2020.

\bibitem{VGGFace}
O.M. Parkhi, A. Vedaldi, and A. Zisserman.
\newblock Deep face recognition.
\newblock In {\em BMVC}, 2015.

\bibitem{PeersSIGGRAPH2007}
Pieter Peers, Naoki Tamura, Wojciech Matusik, and Paul Debevec.
\newblock Post-production facial performance relighting using reflectance
  transfer.
\newblock In {\em SIGGRAPH}, 2007.

\bibitem{facerecognition}
Laiyun Qing, Shiguang Shan, and Xilin Chen.
\newblock Face relighting for face recognition under generic illumination.
\newblock In {\em ICASSP}, 2004.

\bibitem{SfSNet}
Soumyadip Sengupta, Angjoo Kanazawa, Carlos~D. Castillo, and David~W. Jacobs.
\newblock Sf{S}{N}et: {L}earning shape, refectance and illuminance of faces in
  the wild.
\newblock In {\em CVPR}, 2018.

\bibitem{RealisticInverseLighting}
Davoud Shahlaei and Volker Blanz.
\newblock Realistic inverse lighting from a single 2{D} image of a face, taken
  under unknown and complex lighting.
\newblock In {\em FG}, 2015.

\bibitem{ShashuaRatioImage}
Amnon Shashua and Tammy Riklin-Raviv.
\newblock The quotient image: {C}lass-based re-rendering and recognition with
  varying illuminations.
\newblock {\em PAMI}, 2001.

\bibitem{Flickr}
YiChang Shih, Sylvain Paris, Connelly Barnes, William~T. Freeman, and Fredo
  Durand.
\newblock Style transfer for headshot portraits.
\newblock In {\em SIGGRAPH}, 2014.

\bibitem{MassTransport}
Zhixin Shu, Sunil Hadap, Eli Shechtman, Kalyan Sunkavalli, Sylvain Paris, and
  Dimitris Samaras.
\newblock Portrait lighting transfer using a mass transport approach.
\newblock {\em TOG}, 2017.

\bibitem{NeuralFaceEditing}
Zhixin Shu, Ersin Yumer, Sunil Hadap, Kalyan Sunkavalli, Eli Shechtman, and
  Dimitris Samaras.
\newblock Neural face editing with intrinsic image disentangling.
\newblock In {\em CVPR}, 2017.

\bibitem{Stoschek2000}
Arne Stoschek.
\newblock Image-based re-rendering of faces for continuous pose and
  illumination directions.
\newblock In {\em CVPR}, 2000.

\bibitem{UCSDSingleImagePortraitRelighting}
Tiancheng Sun, Jonathan~T Barron, Yun-Ta Tsai, Zexiang Xu, Xueming Yu, Graham
  Fyffe, Christoph Rhemann, Jay Busch, Paul Debevec, and Ravi Ramamoorthi.
\newblock Single image portrait relighting.
\newblock In {\em SIGGRAPH}, 2019.

\bibitem{Tewari2017}
Ayush Tewari, Michael Zollofer, Hyeongwoo Kim, Pablo Garrido, Florian Bernard,
  Patrick Perez, and Theobalt Christian.
\newblock Mofa: Model-based {D}eep {C}onvolutional {F}ace {A}utoencoder for
  {U}nsupervised {M}onocular {R}econstruction.
\newblock In {\em ICCV}, 2017.

\bibitem{towards-high-fidelity-nonlinear-3d-face-morphable-model}
Luan Tran, Feng Liu, and Xiaoming Liu.
\newblock Towards high-fidelity nonlinear 3{D} face morphable model.
\newblock In {\em CVPR}, 2019.

\bibitem{LuanCVPR18}
Luan Tran and Xiaoming Liu.
\newblock Nonlinear 3{D} face morphable model.
\newblock In {\em CVPR}, 2018.

\bibitem{on-learning-3d-face-morphable-model-from-in-the-wild-images}
Luan Tran and Xiaoming Liu.
\newblock On learning 3{D} face morphable model from in-the-wild images.
\newblock {\em PAMI}, 2019.

\bibitem{WangPAMI2009}
Yang Wang, Lei Zhang, Zicheng Liu, Gang Hua, Zhen Wen, Zhengyou Zhang, and
  Dimitris Samaras.
\newblock Face relighting from a single image under arbitrary unknown lighting
  conditions.
\newblock {\em PAMI}, 2009.

\bibitem{SSIM}
Zhou Wang, Alan~C Bovik, Hamid~R Sheikh, and Eero~P Simoncelli.
\newblock Image quality assessment: from error visibility to structural
  similarity.
\newblock {\em TIP}, 2004.

\bibitem{WenRadianceMaps}
Zhen Wen, Zicheng Liu, and Tomas Huang.
\newblock Face {R}elighting with {R}adiance {E}nvironment {M}aps.
\newblock In {\em CVPR}, 2003.

\bibitem{YamaguchiIntrinsic2018}
Shuco Yamaguchi, Shunsuke Saito, Koki Nagano, Yajie Zhao, Weikai Chen, Kyle
  Olszewski, Shigeo Morishima, and Hao Li.
\newblock High-fidelity facial reflectance and geometry inference from an
  unconstrained image.
\newblock In {\em SIGGRAPH}, 2018.

\bibitem{PortraitShadowManipulation}
Xuaner Zhang, Jonathan~T. Barron, Yun-Ta Tsai, Rohit Pandey, Xiuming Zhang, Ren
  Ng, and David~E. Jacobs.
\newblock Portrait {S}hadow {M}anipulation.
\newblock {\em TOG}, 2020.

\bibitem{DPR}
Hao Zhou, Sunil Hadap, Kalyan Sunkavalli, and David~W Jacobs.
\newblock Deep single-image portrait relighting.
\newblock In {\em ICCV}, 2019.

\bibitem{3DDFA}
Xiangyu Zhu, Zhen Lei, Xiaoming Liu, Hailin Shi, and Stan~Z. Li.
\newblock Face {A}lignment {A}cross {L}arge {P}oses: {A} 3{D} {S}olution.
\newblock In {\em CVPR}, 2017.

\bibitem{face-alignment-in-full-pose-range-a-3d-total-solution}
Xiangyu Zhu, Xiaoming Liu, Zhen Lei, and Stan Li.
\newblock Face alignment in full pose range: A 3{D} total solution.
\newblock {\em PAMI}, 2017.

\end{thebibliography}
}

\clearpage
\setcounter{equation}{0}
\setcounter{figure}{0}
\setcounter{table}{0}
\setcounter{section}{0}
\twocolumn[\centering \section*{\Large \textbf{Towards High Fidelity Face Relighting with Realistic Shadows \\ Supplementary Materials\\[3cm]}}] 

%\maketitle
%\thispagestyle{empty}
%\pagestyle{empty}

%------------------------------------------------------------------------
\section{Qualitative Results on FFHQ for Nestmeyer et al. \cite{PhysicsGuidedRelighting}}
%------------------------------------------------------------------------

\renewcommand{\thefigure}{1}
\begin{figure}[t]
\vspace{-2mm}
\begin{center}
   \sidesubfloat[]{\includegraphics[width=0.99 \linewidth]{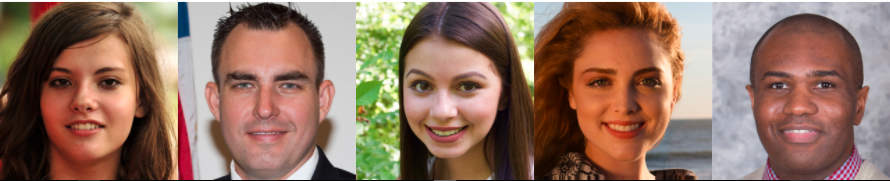}} \\
   \sidesubfloat[]{\includegraphics[width=0.99 \linewidth]{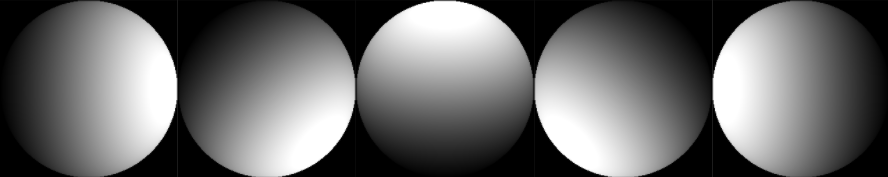}} \\
   \sidesubfloat[]{\includegraphics[width=0.99 \linewidth]{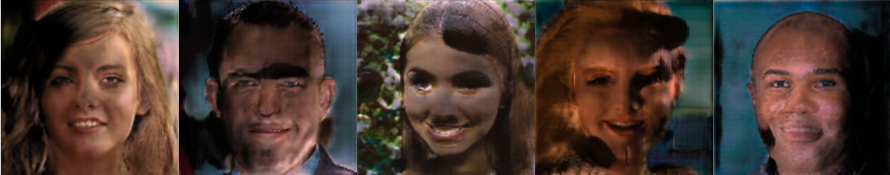}}
\vspace{2mm}
\caption{\textbf{Performance of \cite{PhysicsGuidedRelighting} on FFHQ}. (a) input image, (b) target lighting, (c) Nestmeyer et al. \cite{PhysicsGuidedRelighting}. The method of \cite{PhysicsGuidedRelighting} does not seem to generalize well to in-the-wild images, which is likely caused by the limited subject diversity and night-time setting of their training set. 
}\label{fig:NestmeyerFFHQ}
\end{center}\vspace{-8mm}
\end{figure}

We include qualitative results on FFHQ for Nestmeyer et al. \cite{PhysicsGuidedRelighting} (See Fig.~\ref{fig:NestmeyerFFHQ}). Overall, their model seems to generalize poorly to in-the-wild images, which is likely caused by the limited number of subjects ($21$) in their training set and the night-time setting of their images. Therefore, we chose not to compare with them qualitatively on FFHQ. 

\renewcommand{\thefigure}{2}
\begin{figure*}[t]
\vspace{-2mm}
\begin{center}
\begin{minipage}[t]{0.1375\linewidth}
\centering
\includegraphics[width=\linewidth]{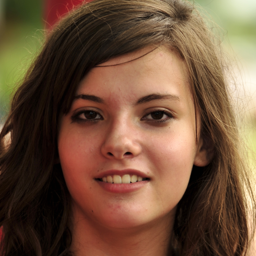}
\end{minipage}
\begin{minipage}[t]{0.1375\linewidth}
\centering
\includegraphics[width=\linewidth]{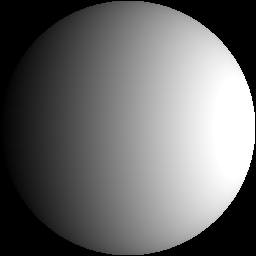}
\end{minipage}
\begin{minipage}[t]{0.1375\linewidth}
\centering
\includegraphics[width=\linewidth]{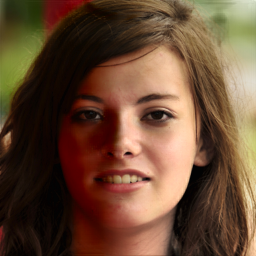}
\end{minipage}
\begin{minipage}[t]{0.1375\linewidth}
\centering
\includegraphics[width=\linewidth]{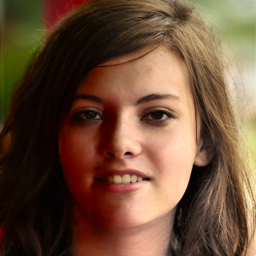}
\end{minipage}
\begin{minipage}[t]{0.1375\linewidth}
\centering
\includegraphics[width=\linewidth]{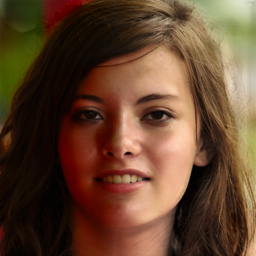}
\end{minipage}
\begin{minipage}[t]{0.1375\linewidth}
\centering
\includegraphics[width=\linewidth]{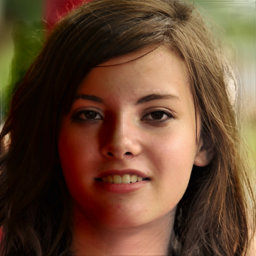}
\end{minipage}
\begin{minipage}[t]{0.1375\linewidth}
\centering
\includegraphics[width=\linewidth]{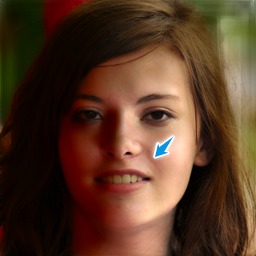}
\end{minipage}

%\newline 

\begin{minipage}[t]{0.1375\linewidth}
\centering
\includegraphics[width=\linewidth]{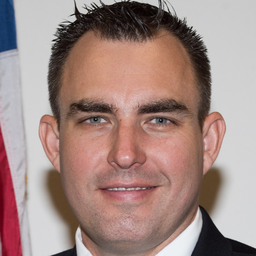}
\end{minipage}
\begin{minipage}[t]{0.1375\linewidth}
\centering
\includegraphics[width=\linewidth]{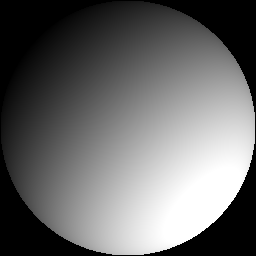}
\end{minipage}
\begin{minipage}[t]{0.1375\linewidth}
\centering
\includegraphics[width=\linewidth]{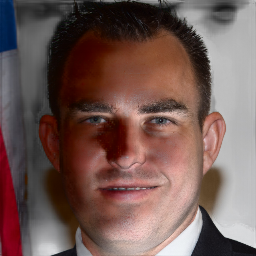}
\end{minipage}
\begin{minipage}[t]{0.1375\linewidth}
\centering
\includegraphics[width=\linewidth]{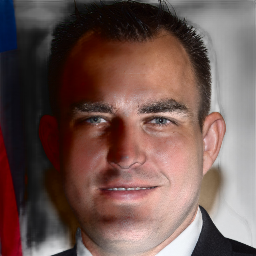}
\end{minipage}
\begin{minipage}[t]{0.1375\linewidth}
\centering
\includegraphics[width=\linewidth]{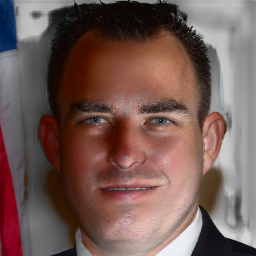}
\end{minipage}
\begin{minipage}[t]{0.1375\linewidth}
\centering
\includegraphics[width=\linewidth]{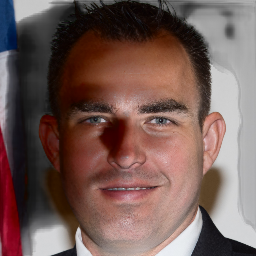}
\end{minipage}
\begin{minipage}[t]{0.1375\linewidth}
\centering
\includegraphics[width=\linewidth]{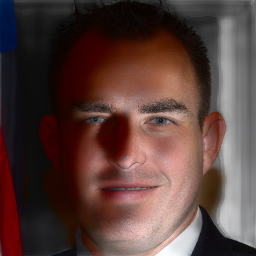}
\end{minipage}

%\newline

\begin{minipage}[t]{0.1375\linewidth}
\centering
\includegraphics[width=\linewidth]{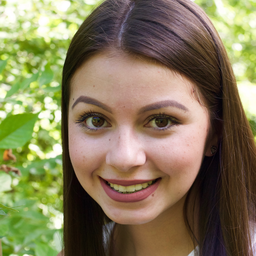}
\end{minipage}
\begin{minipage}[t]{0.1375\linewidth}
\centering
\includegraphics[width=\linewidth]{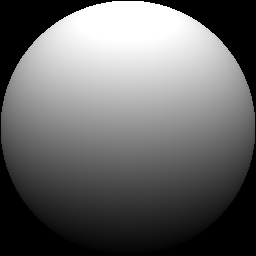}
\end{minipage}
\begin{minipage}[t]{0.1375\linewidth}
\centering
\includegraphics[width=\linewidth]{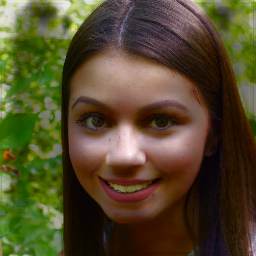}
\end{minipage}
\begin{minipage}[t]{0.1375\linewidth}
\centering
\includegraphics[width=\linewidth]{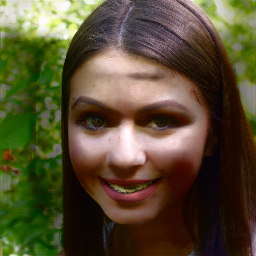}
\end{minipage}
\begin{minipage}[t]{0.1375\linewidth}
\centering
\includegraphics[width=\linewidth]{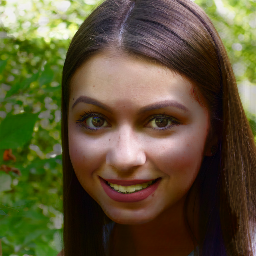}
\end{minipage}
\begin{minipage}[t]{0.1375\linewidth}
\centering
\includegraphics[width=\linewidth]{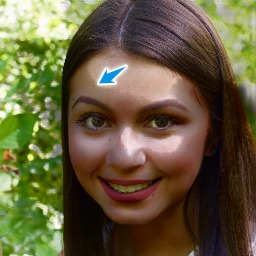}
\end{minipage}
\begin{minipage}[t]{0.1375\linewidth}
\centering
\includegraphics[width=\linewidth]{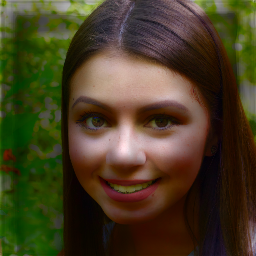}
\end{minipage}

%\newline

\begin{minipage}[t]{0.1375\linewidth}
\centering
\includegraphics[width=\linewidth]{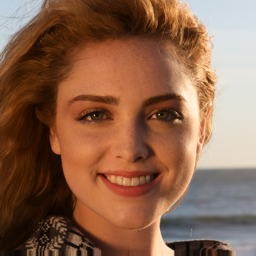}
\end{minipage}
\begin{minipage}[t]{0.1375\linewidth}
\centering
\includegraphics[width=\linewidth]{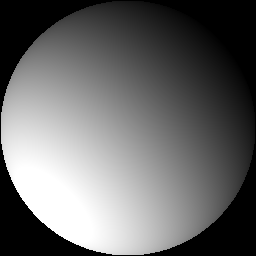}
\end{minipage}
\begin{minipage}[t]{0.1375\linewidth}
\centering
\includegraphics[width=\linewidth]{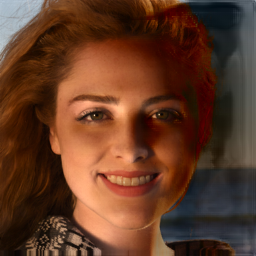}
\end{minipage}
\begin{minipage}[t]{0.1375\linewidth}
\centering
\includegraphics[width=\linewidth]{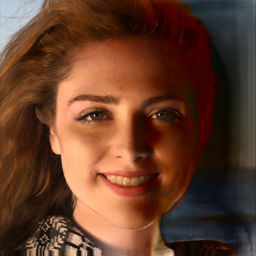}
\end{minipage}
\begin{minipage}[t]{0.1375\linewidth}
\centering
\includegraphics[width=\linewidth]{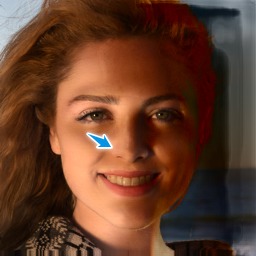}
\end{minipage}
\begin{minipage}[t]{0.1375\linewidth}
\centering
\includegraphics[width=\linewidth]{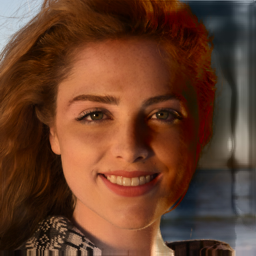}
\end{minipage}
\begin{minipage}[t]{0.1375\linewidth}
\centering
\includegraphics[width=\linewidth]{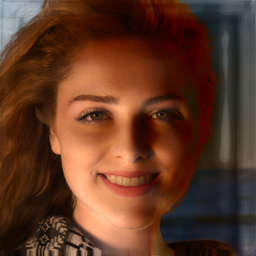}
\end{minipage}

%\newline

\begin{minipage}[t]{0.1375\linewidth}
\centering
\includegraphics[width=\linewidth]{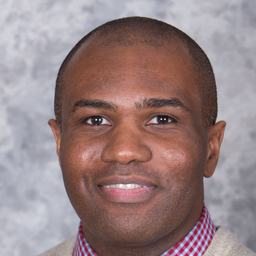}
\small (a) Source Image
\end{minipage}
\begin{minipage}[t]{0.1375\linewidth}
\centering
\includegraphics[width=\linewidth]{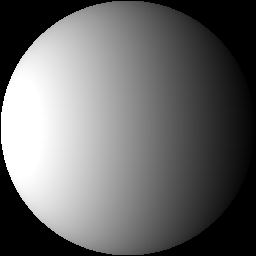}
\small (b) Target Lighting
\end{minipage}
\begin{minipage}[t]{0.1375\linewidth}
\centering
\includegraphics[width=\linewidth]{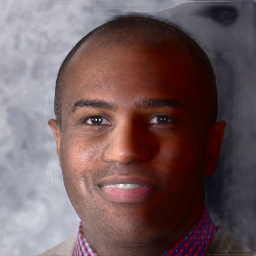}
\small (c) Proposed
\end{minipage}
\begin{minipage}[t]{0.1375\linewidth}
\centering
\includegraphics[width=\linewidth]{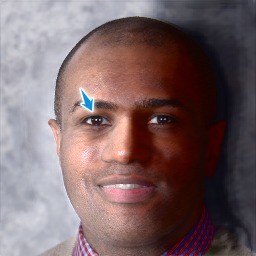}
\small (d) w/o $L_\text{face}$
\end{minipage}
\begin{minipage}[t]{0.1375\linewidth}
\centering
\includegraphics[width=\linewidth]{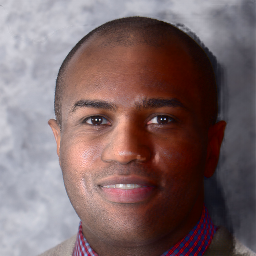}
\small (e) w/o $L_\text{gradient}$
\end{minipage}
\begin{minipage}[t]{0.1375\linewidth}
\centering
\includegraphics[width=\linewidth]{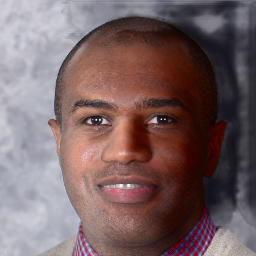}
\small (f) w/o $L_\text{DSSIM}$
\end{minipage}
\begin{minipage}[t]{0.1375\linewidth}
\centering
\includegraphics[width=\linewidth]{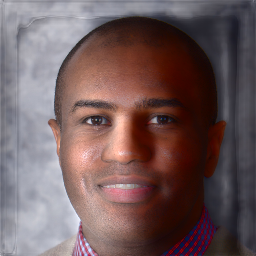}
\small (g) w/o $L_\text{adversarial}$
\end{minipage}
\vspace{-2mm}
\caption{\textbf{Qualitative Results on FFHQ for Ablation Models.} We compare the qualitative performance of our proposed model with $4$ additional ablation models, each of which removes one loss function. Removing $L_\text{face}$ degrades the model's ability to preserve the subject's facial features, removing $L_\text{gradient}$ causes the edges on the face to become smoother (fourth row, around the nose) and lowers the quality of the produced cast shadows, removing $L_\text{DSSIM}$ can lead to unnatural transitions from illuminated regions of the face to shadowed regions (third row), and removing $L_\text{adversarial}$ leads to a noticeably more blurry output. The proposed model that uses all of these losses yields the best qualitative performance. }\label{fig:FFHQAdditionalAblations}
\end{center}\vspace{-2mm}
\end{figure*}

%------------------------------------------------------------------------
\section{Ablations on $L_\text{face}$, $L_\text{gradient}$, $L_\text{DSSIM}$, and $L_\text{adversarial}$}
%------------------------------------------------------------------------

\begin{table}[t!]
\begin{center}
%\small
\setlength\tabcolsep{5.25pt}{  
\begin{tabular}{| c | c | c | c |}
\hline
Method & Si-MSE & MSE & DSSIM \\
\hline
w/o $L_\text{face}$ & $0.0262$ & $0.0346$ & $0.1734$\\
\hline
w/o $L_\text{gradient}$ & $0.0223$ & $0.0297$ & $\mathbf{0.1589}$ \\
\hline
w/o $L_\text{DSSIM}$ & $0.0227$ & $0.0293$ & $0.1703$\\
\hline
w/o $L_\text{adversarial}$ & $0.0275$ & $0.0361$ & $0.1800$ \\
\hline
Proposed & $\mathbf{0.0220}$ & $\mathbf{0.0292}$ & 0.1605 \\
\hline
\end{tabular}
}
\end{center}
\vspace{-3mm}
\caption{
\textbf{Additional Ablation Studies}. We perform ablation studies on $L_\text{face}$, $L_\text{gradient}$, $L_\text{DSSIM}$, and $L_\text{adversarial}$. We find that the proposed model (which includes all loss functions) achieves the best overall performance. 
}\label{tab:AdditionalAblations}
\vspace{-2mm}
\end{table}

We perform $4$ additional ablations to show that the addition of each loss function improves our model's relighting performance. We train $4$ additional models, each of which excludes one of $L_\text{face}$, $L_\text{gradient}$, $L_\text{DSSIM}$, and $L_\text{adversarial}$, and compare the performance with our proposed model. We evaluate their performance quantitatively on Multi-PIE \cite{Multi-PIE} (See Tab.~\ref{tab:AdditionalAblations}). We find that our proposed model, which includes all loss functions, achieves the best overall performance across our $3$ evaluation metrics. 

We further demonstrate the benefits of including these $4$ losses qualitatively on FFHQ \cite{FFHQ} (See Fig.~\ref{fig:FFHQAdditionalAblations}). We find that excluding $L_\text{face}$ lowers the model's ability to preserve the subject's facial details, as expected since $L_\text{face}$ is responsible for ensuring face feature consistency for the same subject under different lighting. Removing $L_\text{gradient}$ can lead to smoother edges on the face, which is understandable given it encourages the edges of the predicted and target ratio images to be similar. Another effect is the quality of the cast shadows generally decreases, which makes sense because shadow borders are edges in the image. Excluding $L_\text{DSSIM}$ can lead to unnatural transitions from illuminated to shadowed regions of the face: rather than transitioning gradually, the shift between the regions can be abrupt. This behavior can arise since the SSIM metric measures if the local image patterns of the predicted image match the target image. Without the corresponding loss $L_\text{DSSIM}$, these unnatural transitions may occur. Finally, we find that removing $L_\text{adversarial}$ noticeably reduces the visual quality of the relit images and makes the output more blurry. This verifies that the inclusion of PatchGAN \cite{PatchGAN} discriminators helps the model capture high frequency details and improve the photorealism of the results. We thus assert that qualitatively, our proposed model produces the best results. 

%------------------------------------------------------------------------
\section{Ratio vs Relit Image Estimation}
%------------------------------------------------------------------------

\renewcommand{\thefigure}{3}
\begin{figure}[t]
\vspace{-2mm}
\begin{center}
\begin{minipage}[t]{0.46\linewidth}
\centering
\includegraphics[width=\linewidth]{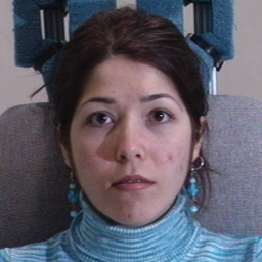}
\end{minipage}
\begin{minipage}[t]{0.46\linewidth}
\centering
\includegraphics[width=\linewidth]{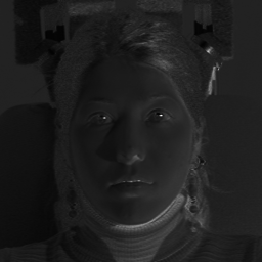}
\end{minipage}

%\newline 

\begin{minipage}[t]{0.46\linewidth}
\centering
\includegraphics[width=\linewidth]{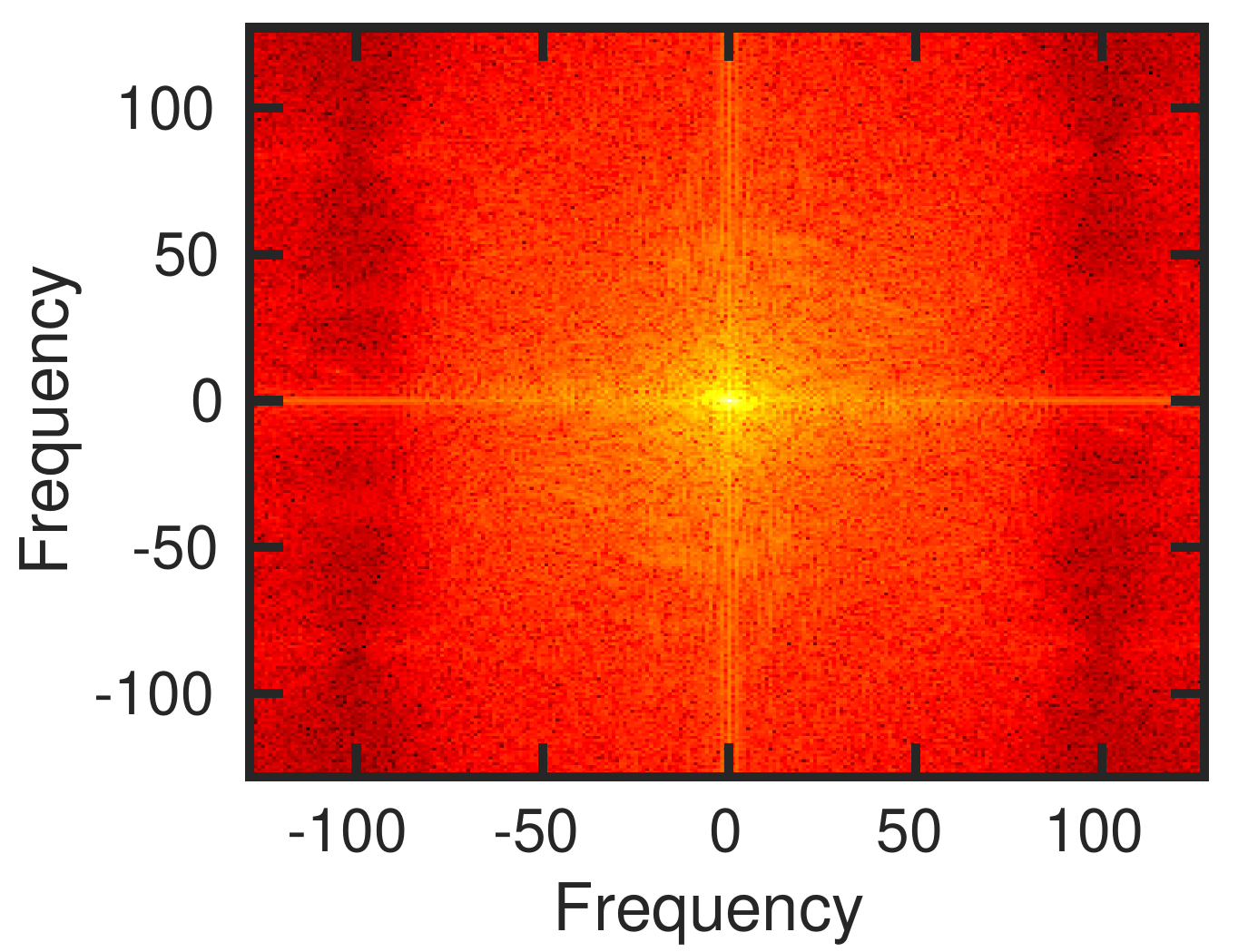}
\small (a) Target Image
\end{minipage}
\begin{minipage}[t]{0.46\linewidth}
\centering
\includegraphics[width=\linewidth]{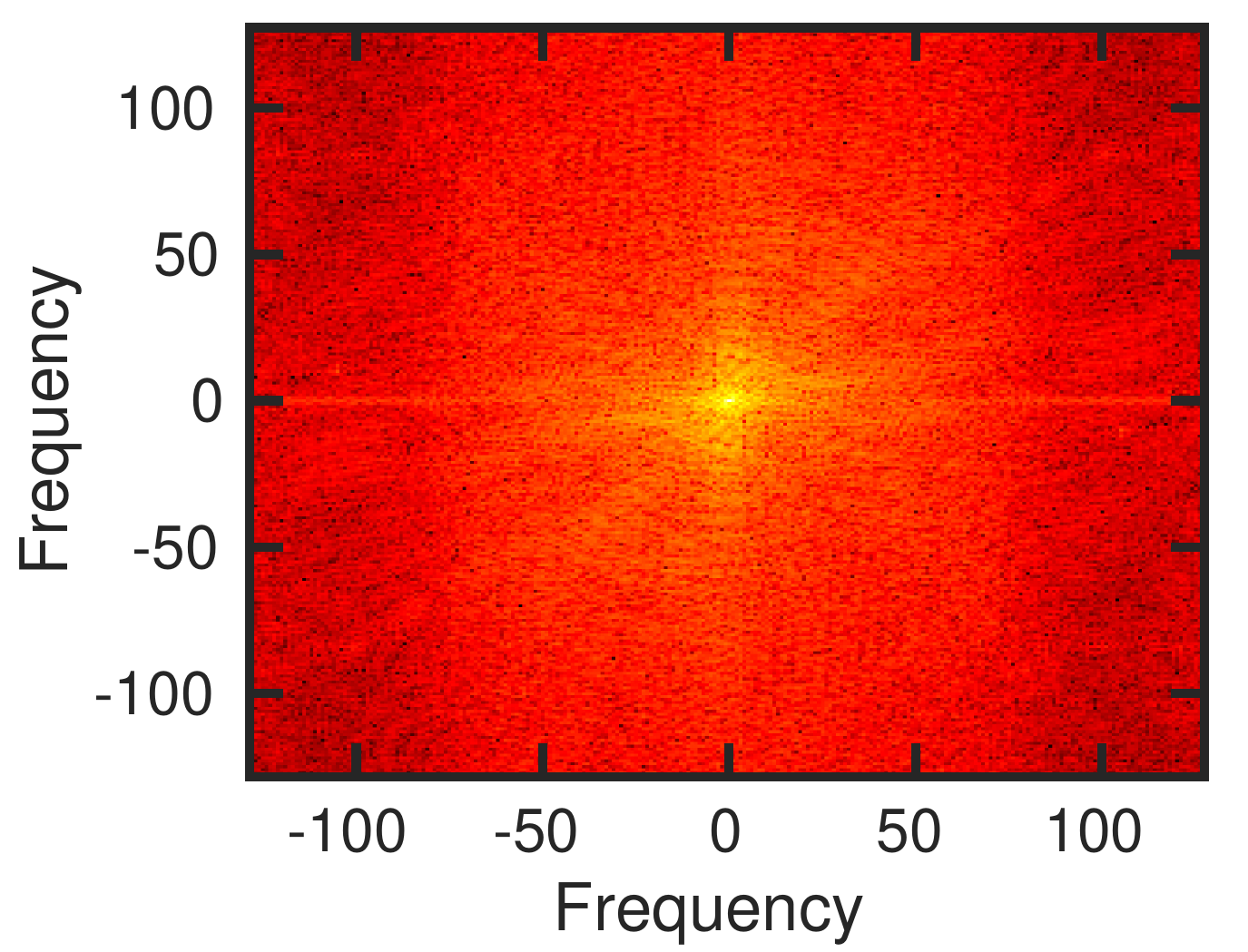}
\small (b) Ratio Image
\end{minipage}
\vspace{-5mm}
\caption{\textbf{Power spectrums}. Comparing the power spectrums of a target Multi-PIE image and its corresponding ratio image, it is clear that the ratio image contains less high-frequency detail than the target image. The ratio image is thus easier for our model to regress than the target image, which indicates that it is an easier way to preserve facial details. 
}\label{fig:PowerSpectrum}
\end{center}
\end{figure}

We provide more insights concerning the advantages of estimating the ratio image instead of the relit image directly. Regressing a ratio image is easier for the network since it contains less high-frequency detail than the target image. This can be shown by the power spectrums of a target Multi-PIE image and its corresponding ratio image in Fig.~\ref{fig:PowerSpectrum}. It's clear from the power spectrums that the target image contains more high-frequency details than the ratio image.
The ratio image contains more lower frequency information, and is therefore easier to estimate. We can thus more easily preserve facial details by regressing ratio images instead of relit images directly.

%------------------------------------------------------------------------
\section{Shadow Removal}
%------------------------------------------------------------------------

\renewcommand{\thefigure}{4}
\begin{figure}[t]
\vspace{-2mm}
\begin{center}
\begin{minipage}[t]{0.32\linewidth}
\centering
\includegraphics[width=\linewidth]{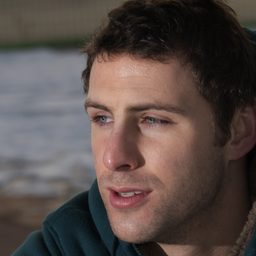}
\end{minipage}
\begin{minipage}[t]{0.32\linewidth}
\centering
\includegraphics[width=\linewidth]{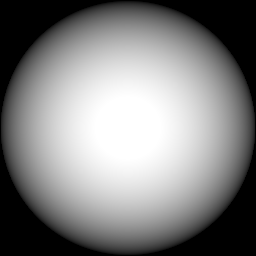}
\end{minipage}
\begin{minipage}[t]{0.32\linewidth}
\centering
\includegraphics[width=\linewidth]{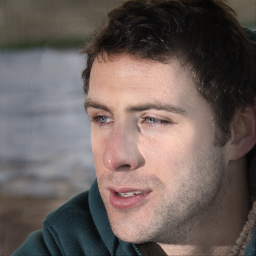}
\end{minipage}

%\newline 

\begin{minipage}[t]{0.32\linewidth}
\centering
\includegraphics[width=\linewidth]{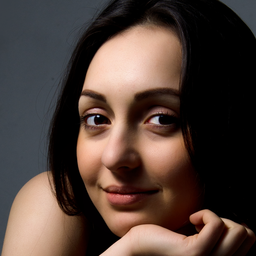}
\small (a) Input Image
\end{minipage}
\begin{minipage}[t]{0.32\linewidth}
\centering
\includegraphics[width=\linewidth]{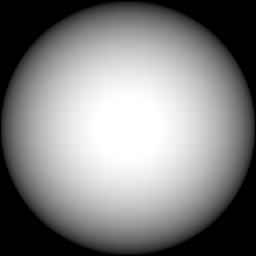}
\small (b) Target Lighting
\end{minipage}
\begin{minipage}[t]{0.32\linewidth}
\centering
\includegraphics[width=\linewidth]{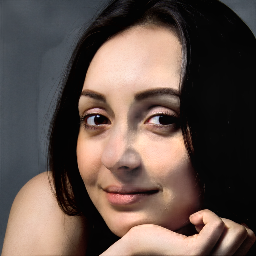}
\small (c) Result
\end{minipage}
   \vspace{-5mm}
\caption{\textbf{Shadow Removal}. Our model is able to remove or significantly soften hard cast shadows. 
}\label{fig:ShadowRemoval}
\end{center}
\end{figure}

As shadow removal is a challenging task in face relighting, we demonstrate our model's capacity to remove hard cast shadows. Fig.~\ref{fig:ShadowRemoval} shows relit results on FFHQ subjects using a frontal target lighting, which should remove facial shadows. Our model either removes hard shadows completely or softens them. 

%------------------------------------------------------------------------
\section{Non-frontal Poses}
%------------------------------------------------------------------------

\renewcommand{\thefigure}{5}
\begin{figure}[t]
\vspace{-2mm}
\begin{center}
\begin{minipage}[t]{0.32\linewidth}
\centering
\includegraphics[width=\linewidth]{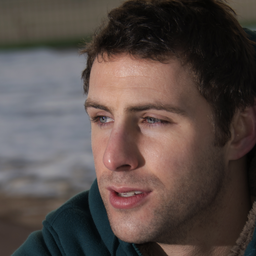}
\end{minipage}
\begin{minipage}[t]{0.32\linewidth}
\centering
\includegraphics[width=\linewidth]{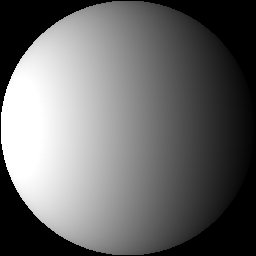}
\end{minipage}
\begin{minipage}[t]{0.32\linewidth}
\centering
\includegraphics[width=\linewidth]{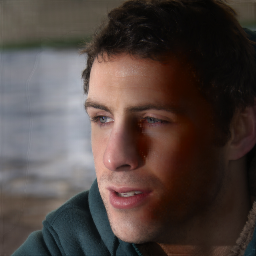}
\end{minipage}

%\newline 

\begin{minipage}[t]{0.32\linewidth}
\centering
\includegraphics[width=\linewidth]{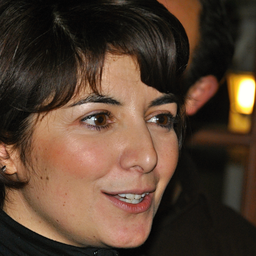}
\small (a) Input Image
\end{minipage}
\begin{minipage}[t]{0.32\linewidth}
\centering
\includegraphics[width=\linewidth]{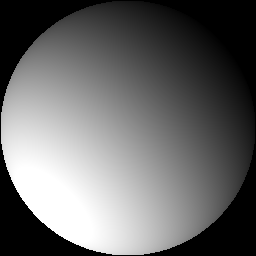}
\small (b) Target Lighting
\end{minipage}
\begin{minipage}[t]{0.32\linewidth}
\centering
\includegraphics[width=\linewidth]{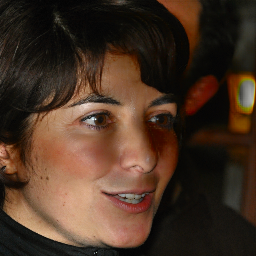}
\small (c) Result
\end{minipage}
    \vspace{-5mm}
\caption{\textbf{Non-frontal poses}. Our model is able to properly relight images with large, non-frontal poses. 
}\label{fig:NonFrontal}
\end{center}
\end{figure}

To demonstrate our model's performance on face images with large poses, we apply our model to FFHQ subjects with non-frontal poses. As shown in Fig.~\ref{fig:NonFrontal}, our model can handle large poses gracefully.  

%------------------------------------------------------------------------
\section{Lighting Estimation}
%------------------------------------------------------------------------

We provide more details on how we estimate the groundtruth lightings for the DPR \cite{DPR}, Yale \cite{ExtendedYale}, and Multi-PIE \cite{Multi-PIE} datasets. 

The DPR images provide the SH coefficients as groundtruth lighting. The Yale dataset provides the lighting direction for each image, and we treat each light as a directional light. Multi-PIE provides positions for every light, which we treat as point lights. For Yale and Multi-PIE, we project the light source to SH basis functions on the unit sphere to get the SH coefficients. The ambient component is estimated using shadow masks, as explained in Sec. 3.5 of the main paper. 

%------------------------------------------------------------------------
\section{Inference Time} 
%------------------------------------------------------------------------

We compare our model's inference time with SfSNet \cite{SfSNet} and Nestmeyer \textit{et al.} \cite{PhysicsGuidedRelighting}, two methods with available code and significantly different architectures from our work. Running on Multi-PIE, \cite{SfSNet} takes $0.0087$ seconds per image, \cite{PhysicsGuidedRelighting} takes $0.0750$ seconds per image, and our model is the fastest at $0.0075$ seconds per image. 

%------------------------------------------------------------------------
\section{Qualitative Video}
%------------------------------------------------------------------------

We include a video with $5$ FFHQ subjects to show that our model can faithfully perform face relighting across many different target lighting directions and many diverse subjects. Our video also shows that our model can handle varying lighting intensities, as seen when we move the light source closer to the last subject at the end of the video. 

%------------------------------------------------------------------------
\end{document}